\documentclass[10pt,twocolumn]{article}

\usepackage{cvpr}
\usepackage{times}
\usepackage{epsfig}
\usepackage{graphicx}
\usepackage{amsmath}
\usepackage{amssymb}
\usepackage{stfloats}
\usepackage{tablefootnote}

\usepackage[dvipsnames]{xcolor} % enable colorful writing
\usepackage{float}
\usepackage{caption} % solve for table caption spacing 
\usepackage[section]{placeins}
\usepackage{subcaption}
\usepackage{multirow} % for tabular multi-row
\usepackage{enumitem}
\newcommand{\denselist}{\itemsep -2pt\parsep=1pt\partopsep 0pt}
\newcommand{\bitem}{\begin{itemize}\denselist}
\newcommand{\eitem}{\end{itemize}}
\newcommand{\benum}{\begin{enumerate}\denselist}
\newcommand{\eenum}{\end{enumerate}}

\newcommand{\Noindent}{\vspace{2pt} \noindent}
\newcommand{\partitle}[1]{\Noindent\textbf{#1: }}
\newcommand{\parquest}[1]{\Noindent\textbf{#1?\\}}

\newcommand{\ngap}{\quad\quad}
\newcommand{\name}{SIGNet\xspace}

\usepackage[pagebackref=true,breaklinks=true,colorlinks,bookmarks=false]{hyperref}
\cvprfinalcopy % *** Uncomment this line for the final submission and arxiv submission
\def\cvprPaperID{5263} % *** Enter the CVPR Paper ID here

\ifcvprfinal\fi

\begin{document}

%%%%%%%%% TITLE
\title{\name: Semantic Instance Aided Unsupervised 3D Geometry Perception}

%%%%%%%%% AUTHORS
\author{Yue Meng$^1$ \ngap Yongxi Lu$^1$\ngap Aman Raj$^1$\ngap Samuel Sunarjo$^1$\ngap Rui Guo$^2$\\Tara Javidi$^1$\ngap Gaurav Bansal$^2$\ngap Dinesh Bharadia$^1$\\ \\
$^1$UC San Diego \ngap $^2$Toyota InfoTechnology Center\\
{\tt\small \{yum107, yol070, amraj, ssunarjo, tjavidi, dineshb\}@ucsd.edu} \\ 
{\tt\small rguo@us.toyota-itc.com\ngap gauravbs@gmail.com}}

\maketitle
\thispagestyle{empty}
%%%%%%%%% ABSTRACT
\begin{abstract}
Unsupervised learning for geometric perception (depth, optical flow, etc.) is of great interest to autonomous systems. Recent works on unsupervised learning have made considerable progress on perceiving geometry; however, they usually ignore the coherence of objects and perform poorly under scenarios with dark and noisy environments. In contrast, supervised learning algorithms, which are robust, require large labeled geometric dataset. This paper introduces SIGNet, a novel framework that provides robust geometry perception without requiring  geometrically informative labels. Specifically, SIGNet integrates semantic information to make depth and flow predictions consistent with objects and robust to low lighting conditions. SIGNet is shown to improve upon the state-of-the-art unsupervised learning for depth prediction by 30\% (in squared relative error). In particular, SIGNet improves the dynamic object class performance by 39\% in depth prediction and 29\% in flow prediction. Our code will be made available at \url{https://github.com/mengyuest/SIGNet}
\end{abstract}

%%%%%%%%% BODY
\section{Introduction}
\label{sec:intro}

Visual perception of 3D scene geometry using a monocular camera is a fundamental problem with numerous applications, like autonomous driving and space exploration. We focus on the ability to infer accurate geometry (depth and flow) of static and moving objects in a 3D scene. Supervised deep learning models have been proposed for geometry predictions, yielding ``robust" and favorable results against the traditional approaches (SfM) \cite{saxena2006learning,saxena2009make3d,fischer2015flownet,byravan2017se3,bloesch2018codeslam,lianos2018vso}. However, supervised models require a dataset labeled with geometrically informative annotations, which is extremely challenging as the collection of geometrically annotated ground truth (e.g. depth, flow) requires expensive equipment (e.g. LIDAR) and careful calibration procedures. 

Recent works combine the geometric-based SfM methods with end-to-end unsupervised trainable deep models to utilize abundantly available unlabeled monocular camera data. In \cite{zhou2017unsupervised,vijayanarasimhan2017sfm,yin2018geonet,fei2018geo} deep models predict depth and flow per pixel simultaneously from a short sequence of images and typically use photo-metric reconstruction loss of a target scene from neighboring scenes as the surrogate task. However, these solutions often fail when dealing with dynamic objects\footnote{Section~\ref{sec:experiments} presents empirical results that 
explicitly illustrate this shortcoming of state-of-the-art unsupervised approaches.}.
 \textit{Furthermore, the prediction quality is negatively affected by the imperfections like Lambertian reflectance and varying intensity which occur in the real world. In short, no robust solution is known}.  

\begin{figure}[!t]
    \centering
    \includegraphics[width=\linewidth, height=.7\linewidth]{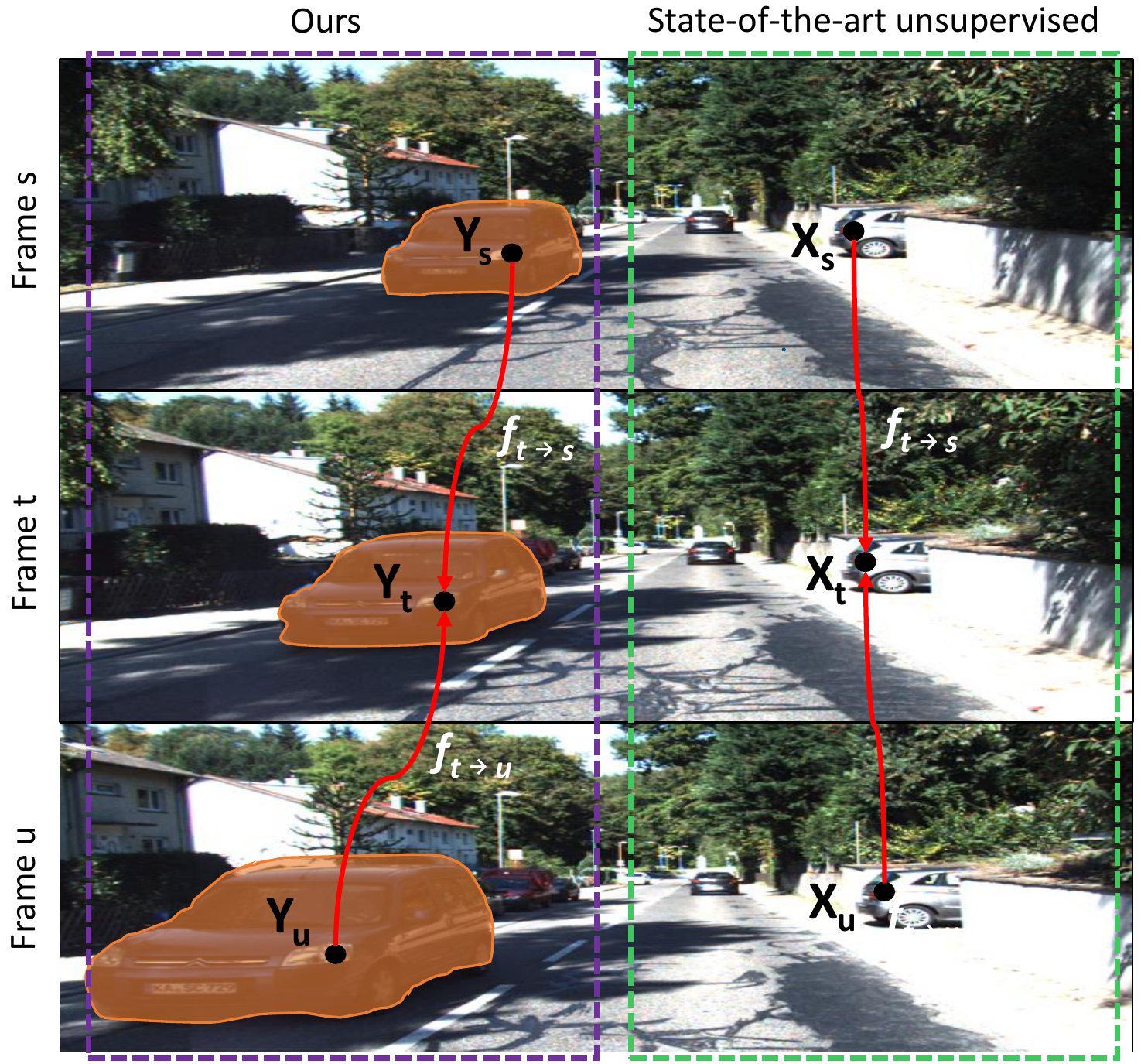}
    \caption{On the right, state-of-the-art unsupervised learning approach relies on pixel-wise information only, while \name on the left utilizes the semantic information to encode the spatial constraints hence further enhances the geometry prediction.}
    \label{fig:translation_across_frames}
\end{figure}

In Fig~\ref{fig:translation_across_frames}, we highlight the innovation of our system (on the left) comparing to the existing unsupervised frameworks (on the right) for geometry perception. Traditional unsupervised models learn from the pixel-level feedback (i.e. photo-metric reconstruction loss), whereas \name relies on the key observation that inherent spatial constraints exist in the visual perception problem as shown in Fig~\ref{fig:translation_across_frames}. Specifically, we exploit the fact that pixels belonging to the same object have additional constraints for the depth and flow prediction.

How can those spatial constraints of the pixels be encoded? We leverage the semantic information as seen in Fig~\ref{fig:translation_across_frames} for unsupervised frameworks. Intuitively, semantic information can be interpreted as defining boundaries around a group of pixels whose geometry is closely related. The knowledge of semantic information between different segments of a scene could allow us to easily learn which pixels are correlated, while the object edges could imply sharp depth transition. Furthermore, note that this learning paradigm is practical~\footnote{Semantic labels can be easily curated on demand  on unlabeled data. On the contrary, geometrically informative labels such as flow and depth require additional sensors and careful annotation at the data collection stage.} as annotations for semantic prediction tasks such as semantic segmentation are relatively cheaper and easier to acquire. To the best of our knowledge, our work is the first to utilize semantic information in the context of unsupervised learning for geometry perception. 

A natural question is how do we combine semantic information with an unsupervised geometric prediction? Our approach to combine the semantic information with RGB input is two-fold: First, we propose a novel way to augment RGB images with semantic information. Second, we propose new loss functions, architecture, and training method. The two-fold approach precisely
accounts for spatial constraints in making geometric predictions:

\partitle{Feature Augmentation}We concatenate the RGB input data with both per-pixel class predictions and instance-level predictions. We use per pixel class predictions to define semantic mask which serves as a guidance signal that eases unsupervised geometric predictions. Moreover, we use the instance-level prediction and split them into two inputs, instance edges and object masks. Instance edges and object masks enable the network to learn the object edges and sharp depth transitions.

\partitle{Loss Function Augmentation}Second, we augment the loss function to include various semantic losses, which reduces the reliance on semantic features in the evaluation phase. This is crucial when the environment contains less common contextual elements (like in dessert navigation or mining exploitation). We design and experiment with various semantic losses, such as semantic warp loss, masked reconstruction loss, and semantic-aware edge smoothness loss. However, manually designing a loss term which can improve the performance over the feature augmentation technique turns out to be very difficult. The challenge comes from the lack of understanding of error distributions because we are generally biased towards simple, interpretable loss functions that can be sub-optimal in unsupervised learning. Hence, we propose an alternative approach of incorporating a transfer network that learns how to predict semantic mask via a semantic reconstruction loss and provides feedback to improve the depth and pose estimations, which shows considerable improvements in depth and flow prediction.

We empirically evaluate the feature and loss function augmentations on KITTI dataset \cite{geiger2012we} and compare them with the state-of-the-art unsupervised learning framework~\cite{yin2018geonet}. In our experiments we use class-level predictions from DeepLabv3+ \cite{chen2018encoder} trained on Cityscapes \cite{cordts2016cityscapes} and Mask R-CNN \cite{he2017mask} trained on MSCOCO \cite{lin2014microsoft}. Our key findings:
\bitem
\item By using semantic segmentation for both feature and loss augmentation, our proposed algorithms improves squared relative error in depth estimation by $28$\% compared to the strong baseline set by state-of-the-art unsupervised GeoNet \cite{yin2018geonet}.

\item Feature augmentation alone, combining semantic with instance-level information, leads to larger gains. With both class-level and instance-level features, the squared relative error of the depth predictions improves by $30$\% compared to the baseline. 

\item Finally, as for common dynamic object classes (e.g. vehicles) \name shows $39$\% improvement (in squared relative error) for depth predictions and $29$\% improvement in the flow prediction, thereby showing that semantic information is very useful for improving the performance in the dynamic categories of objects. Furthermore, \name is robust to noise in image intensity compared to the baseline.\\

\eitem
\section{Related Work}

\partitle{Deep Models for Understanding Geometry}Deep models have been widely used in supervised depth estimation \cite{eigen2014depth, liu2016learning, MIFDB16, Zhang2015MonocularOI, Chen:2016:SDP:3157096.3157178, Xu_2017_CVPR, Xu_2018_CVPR, Fu_2018_CVPR, Xian_2018_CVPR}, tracking, and pose estimation \cite{wang2015visual, xiang2017posecnn, byravan2017se3, gordon2018re} , as well as optical flow predictions \cite{Dosovitskiy_2015_ICCV, Ilg_2017_CVPR, Lai2017SemiSupervisedLF, Sun_2018_CVPR}. These models have demonstrated superior accuracy and typically faster speed in modern hardware platforms (especially in the case of optical flow estimation) compared to traditional methods. However, achieving good performance with supervised learning requires a large amount of geometry-related labels. The current work addresses this challenge by adopting an unsupervised learning framework for depth, pose, and optical flow estimations.

\partitle{Deep Models for Semantic Predictions}Deep models are widely applied in semantic prediction tasks, such as image classification \cite{Krizhevsky:2017:ICD:3098997.3065386}, semantic segmentation \cite{chen2018encoder}, and instance segmentation \cite{he2017mask}. In this work, we utilize the effectiveness of the semantic predictions provided by DeepLab v3+ \cite{chen2018encoder} and Mask R-CNN \cite{he2017mask} in encoding spatial constraints to accurately predict geometric attributes such as depth and flow. While we particularly choose  \cite{chen2018encoder} and  \cite{he2017mask} for our \name,
similar gains can be obtained by using other state-of-the-art semantic prediction methods.

\partitle{Unsupervised Deep Models for Understanding Geometry}Several recent methods propose to use unsupervised learning for geometry understanding. In particular, Garg \etal \cite{Garg2016UnsupervisedCF} uses a warping method based on Taylor expansion. In the context of unsupervised flow prediction, Yu \etal \cite{jason2016back} and Ren \etal \cite{ren2017unsupervised} introduce image reconstruction loss with spatial smoothness constraints. Similar methods are used in Zhou \etal \cite{zhou2017unsupervised} for learning depth and camera ego-motions by ignoring object motions. This is partially addressed by Vijayanarasimhan \etal \cite{vijayanarasimhan2017sfm}, despite the fact, we note, that the modeling of motion is difficult without introducing semantic information. This framework is further improved with better modeling of the geometry. Geometric consistency loss is introduced to handle occluded regions, in binocular depth learning \cite{godard2017unsupervised}, flow prediction \cite{Meister:2018:UUL} and joint depth, ego-motion and optical flow learning \cite{yin2018geonet}. Mahjourian \etal \cite{Mahjourian_2018_CVPR} focuses on improved geometric constraints, Godard \etal \cite{Godard2018DiggingIS} proposes several architectural and loss innovations, while Zhan \etal \cite{Zhan_2018_CVPR} uses reconstruction in the feature space rather than the image space. In contrast, the current work explores using semantic information to resolve ambiguities that are difficult for pure geometric modeling. Methods proposed in the current work are complementary to these recent methods, but we choose to validate our approach on a state-of-the-art framework known as GeoNet \cite{yin2018geonet}.

\partitle{Multi-Task Learning for Semantic and Depth}Multi-task learning \cite{Caruana:1997:ML:262868.262872} achieves better generalization by allowing the system to learn features that are robust across different tasks. Recent methods focus on designing efficient architectures that can predict related tasks using shared features while avoiding negative transfers \cite{Misra2016CrossStitchNF,He:2017:AWM:3123266.3123424,Lu_2017_CVPR,Meyerson2017BeyondSH,8100062,Gao2018NDDRCNNLF}. In this context, several prior works report promising results combining scene geometry with semantics. For instance, similar to our method Liu \etal \cite{Liu2010SingleID} uses semantic predictions to provide depth. However, this work is fully supervised and only uses sub-optimal traditional methods. Wang \etal \cite{Wang_2015_CVPR}, Cross-Stitching \cite{Misra2016CrossStitchNF}, UberNet \cite{8100062} and NDDR-CNN \cite{Gao2018NDDRCNNLF} all report improved performance over single-task baselines. But they have not addressed outdoor scenes and unsupervised geometry understanding. Our work is also related to PAD-Net \cite{Xu2018PADNetMG}. PAD-Net reports improvements by combining intermediate tasks as inputs to final depth and segmentation tasks. Our method of using semantic input similarly introduces an intermediate prediction task as input to the depth and pose predictions, but we tackle the problem setting where depth labels are not provided. 
\section{State-of-the-art Unsupervised Geometry Prediction}
\label{sec:background}

\begin{figure*}
    \centering
    \includegraphics[width=\linewidth]{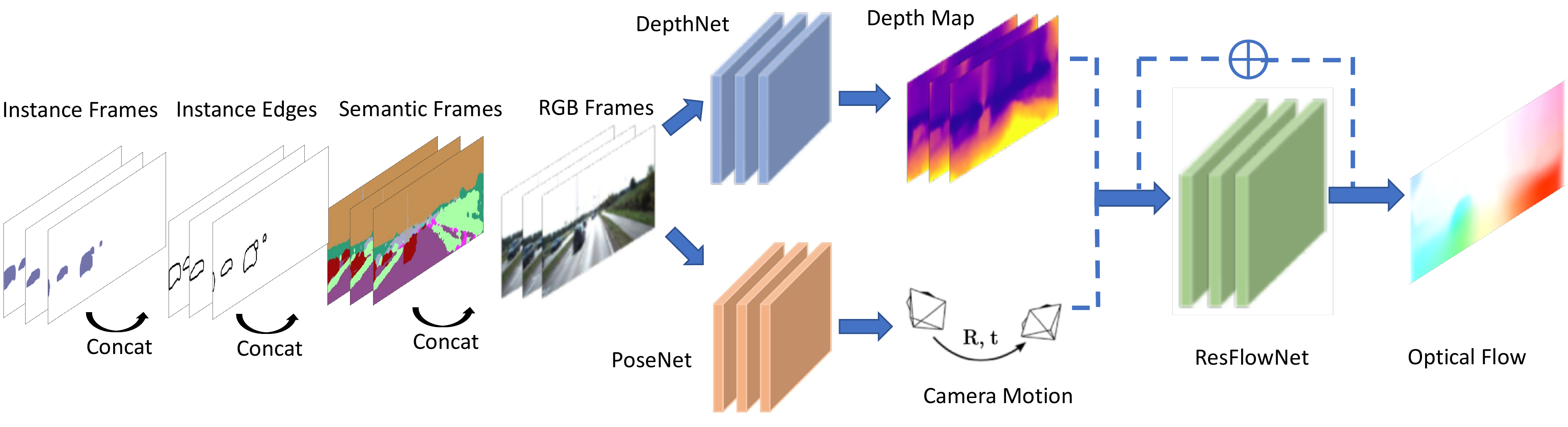}
    \caption{Our unsupervised architecture contains DepthNet, PoseNet and ResFlowNet to predict depth, poses and motion using semantic-level and instance-level segmentation concatenated along the input channel dimension. }
    \label{fig:arch_with_semantic_input}
\end{figure*}

Prior to presenting our technical approach, we provide a brief overview of state-of-the-art unsupervised depth and motion estimation framework, which is based on image reconstruction from geometric predictions \cite{zhou2017unsupervised,yin2018geonet}. It trains the geometric prediction models through the reconstructions of a target image from source images. The target and source images are neighboring frames in a video sequence. Note that such a reconstruction is possible only when certain elements of the 3D geometry of the scene are understood: (1) The relative 3D location (and thus the distance) between the camera and each pixel. (2) The camera ego-motion. (3) The motion of pixels. Thus this framework can be used to train a depth estimator and an ego-motion estimator, as well as a optical flow predictor.

Technically, each training sample $\mathcal{I}=\{I_i\}_{i=1}^n$ consists of $n$ contiguous video frames $I_i\in \mathbb{R}^{H\times W\times 3}$ where the center frame $I_t$ is the ``target frame'' and the other frames serve as the ``source frame''. In training, a differentiable warping function $f_{t\to s}$ is constructed from the geometry predictions. The warping function is used to reconstruct the target frame $\tilde{I}_s \in \mathbb{R}^{H\times W\times 3}$ from source frame $I_s$ via bilinear sampling. The level of success in this reconstruction provides training signals through backpropagation to the various ConvNets in the system. A standard loss function to measure reconstruction success is as follows:
\begin{equation}
\label{eqn:image_simlarity}
\mathcal{L}_{rw}=\alpha \frac{1-\text{SSIM}(I_t,\tilde{I}_s)}{2} + (1-\alpha) ||I_t-\tilde{I}_s||_1
\end{equation}
where SSIM denotes the structural similarity index \cite{wang2004image} and $\alpha$ is set to $0.85$ in \cite{yin2018geonet}.

To filter out erroneous predictions while preserving sharp details, the standard practice is to include an edge-aware depth smoothness loss $\mathcal{L}_{ds}$ weighted by image gradients
\begin{equation}
\label{eqn:edge_aware_smothness}
    \mathcal{L}_{ds}=\sum\limits_{p_t}|\nabla D(p_t)|\cdot (e^{-|\nabla I(p_t)|})^T
\end{equation}
where $|\cdot|$ denotes element-wise absolute operation, $\nabla$ is the vector differential operator, and $T$ denotes transpose of gradients. These losses are usually computed from a pyramid of multi-scale predictions. The sum is used as the training target.

While the reconstruction of RGB images is an effective surrogate task for unsupervised learning, it is limited by the lack of semantic information as supervision signals. For example, the system cannot learn the difference between the car and the road if they have similar colors or two neighboring cars with similar colors. When object motion is considered in the models, the learning can mistakenly assign motion to non-moving objects as the geometric constraints are ill-posed. We augment and improve this system by leveraging semantic information.
\section{Methods}
In this section, we present solutions to enhance geometry predictions with semantic information. Semantic labels can provide rich information on 3D scene geometry. Important details such as 3D location of pixels and their movements can be inferred from a dense representation of the scene semantics. The proposed methods are applicable to a wide variety of recently proposed unsupervised geometry learning frameworks based on photometric reconstruction \cite{zhou2017unsupervised, godard2017unsupervised, yin2018geonet} represented by our baseline framework introduced in Section \ref{sec:background}. Our complemented pipeline in test time is illustrated in Fig \ref{fig:arch_with_semantic_input}.

\begin{figure}
    \centering
    \includegraphics[width=\linewidth]{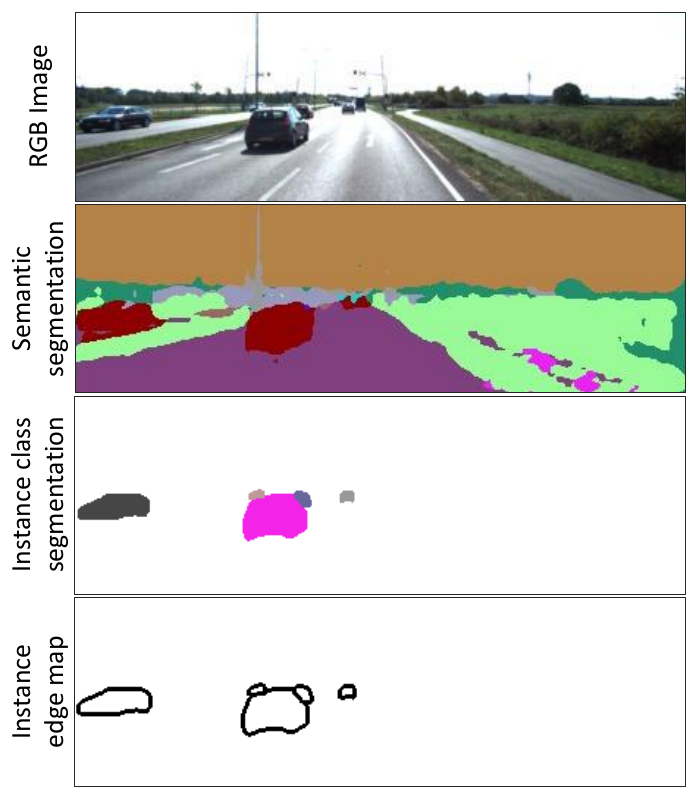}
    \caption{Top to bottom: RGB image, semantic segmentation,  instance class segmentation and instance edge map. They are used for the full prediction architecture. The semantic segmentation provides accurate segments grouped by classes, but it fails to differentiate neighboring cars. }
    \label{fig:sem_side_by_side}
\end{figure}

\subsection{Semantic Input Augmentation}
\label{sec:sem_input}
Semantic predictions can improve geometry prediction models when serving as input features. Unlike RGB images, semantic predictions mark objects and contiguous structures with consistent blobs, which provide important information for the learning problem. However, it is uncertain that using semantic labels as input could indeed improve depth and flow predictions since training labels are not available. Semantic information could be lost or distorted, which would end up being a noisy training signal. An important finding of our work is that using semantic predictions as inputs significantly improves the accuracy in geometry predictions, despite the presence of noisy training signal. Input representation and the type of semantic labels have a large impact on the performance of the system. We further illustrate this by Fig \ref{fig:sem_side_by_side}, where we show various semantic labels (semantic segmentation, instance segmentation, and instance edge) that we use to augment the input. This imposes additional constraints such as depth of the pixels belonging to a particular object (e.g. a vehicle) which helps the learning process. Furthermore, sudden changes in the depth predictions can be inferred from the boundary of vehicles. The semantic labels of the pixels can provide important information to associate pixels across frames.

\partitle{Encoding Pixel-wise Class Labels}We explored two input encoding techniques for class labels: dense encoding and one-hot encoding. In dense encoding, dense class labels are concatenated along the input channel dimension. The added semantic features are centralized to the range of $[-1, 1]$ to be consistent with RGB inputs.  In the case of one-hot encoding, the class-level semantic predictions are first expanded to one-hot encoding and then concatenated along the input channel dimension. The labels are represented as one-hot sparse vectors. In this variant, semantic features are not normalized since they have similar value range as the RGB inputs, 

\partitle{Encoding Instance-level Semantic Information}Both dense and one-hot encoding are natural to class-level semantic prediction, where each pixel is only assigned a class label rather than an instance label. Our conjecture is that instance-level semantic information is particular well-suited to improve unsupervised geometric predictions, as it provides accurate information on the boundary between individual objects of the same type. Unlike class-level label, the instance label itself does not have a well-defined meaning. Across different frames, the same label could refer to different object instances. To efficiently represent the instance-level information, we compute the gradient map of a dense instance map and use it as an additional feature channel concatenating to the class label input (dense/one-hot encoding).

\partitle{Direct Input versus Residual Correction}Complementary to the choice of encoding, we also experiment with different architectures to feed semantic information to the geometry prediction model. In particular, we make a residual prediction using a separate branch that takes in only semantic inputs. Notably, using residual depth prediction leads to further improvement on top of the gains from the direct input methods.

\subsection{Semantic Guided Loss Functions}
\label{method:semantic_loss}
The information from semantic predictions could be diminished due to noisy semantic labels and very deep architectures. Hence, we design training loss functions that are guided by semantic information. In such design, the semantic predictions provide additional loss constraints to the network. In this subsection, we introduce a set of semantic guided loss functions to improve depth and flow predictions. 

\partitle{Semantic Warp Loss}Semantic predictions can help learn scenarios where reconstruction of the RGB image is correct in terms of pixel values but violates obvious semantic correspondences, e.g. matching pixels to incorrect semantic classes and/or instances. In light of this, we propose to reconstruct the semantic predictions in addition of doing so for RGB images. We call this ``semantic warping loss'' as it is based on warping of the semantic predictions from source frames to the target frame. Let $S_s$ be the source frame semantic prediction and $\tilde{S}_s^{rig}$ be the warped semantic image, we define semantic warp loss as:

\begin{equation}
    \mathcal{L}_{sem}=||\tilde{S}_s^{rig}-S_t||_2
\end{equation}
The warped loss is added to the baseline framework using a hyper-tuned value of the weight $w$.

\partitle{Masking of Reconstruction Loss via Semantics}As described in Section \ref{sec:background}, the ambiguity in object motion can lead to sub-optimal learning. Semantic labels can partially resolve this by separating each class of region. Motivated by this observation, we mask the foreground region out to form a set of new images $J_{t,c}^k=I_{t,c}\odot S_{t,k}$ for $c=0,1,2$ and $k=0,...,K-1$ where $c$ represents the RGB-channel index, $\odot$ is the element-wise multiplication operator and $S_{s,k}$ is the $k$-th channel of the binary semantic segmentation ($K$ classes in total). Similarly we can obtain $\tilde{J}_{s,c}^{rig,k}=\tilde{I}_{s,c}^{rig}\odot S_{t,k}$ for $c=0,1,2$ and $k=0,...,K-1$. Finally, the image similarity loss is defined as:
    \begin{equation}
        \mathcal{L}_{rw}'=\sum\limits_{k=0}^{K-1}\alpha \frac{1-\text{SSIM}(J_t^k,\tilde{J}_s^{rig,k})}{2} 
         + (1-\alpha) ||J_t^k-\tilde{J}_s^{rig,k}||_1
    \end{equation}

\begin{figure}[ht]
    \centering
    \includegraphics[width=\linewidth]{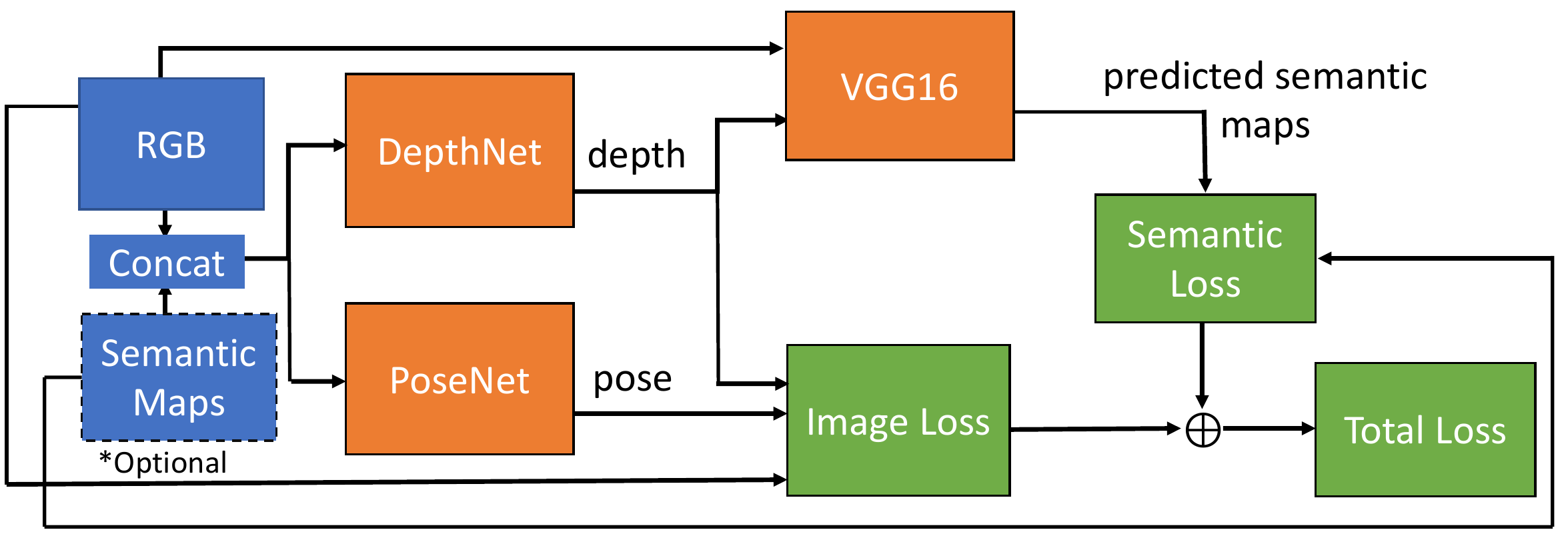}
    \caption{Infer semantic labels from depth predictions. The transfer function uses RGB and predicted depth as input. We experimented the variants with and without semantic input. }
    \label{fig:arch_with_transfer}
\end{figure}

\partitle{Semantic-Aware Edge Smoothness Loss}Equation \ref{eqn:edge_aware_smothness} uses RGB to infer edge locations when enforcing smooth regions of depth. This could be improved by including an edge map computed from semantic predictions. Given a semantic segmentation result $S_{t}$, we define a weight matrix $M_{t} \in [0,1]^{H\times W}$ where the weight is low (close to zero) on class boundary regions and high (close to one) on other regions. We propose a new image similarity loss as:
    \begin{equation}
    \begin{aligned}
    \mathcal{L}_{rw}''&=\sum\limits_{k=0}^{K-1}\alpha \frac{1-\text{SSIM}(I_t\odot M_t,\tilde{I}_s^{rig}\odot M_t)}{2} \\
    & + (1-\alpha) ||I_t\odot M_t-\tilde{I}_s^{rig}\odot M_t||_1
    \end{aligned}
    \end{equation}

\partitle{Semantic Loss by Transfer Network}Motivated by the observation that high-quality depth maps usually depict object classes and background region, we designed \textit{a novel transfer network architecture}. As shown in Fig \ref{fig:arch_with_transfer} the transfer network block receives predicted depth maps along with the original RGB images and outputs semantic labels. The transfer network introduces a semantic reconstruction loss term to the objective function to force the predicted depth maps to be richer in contextual sense, hence refines the depth estimation. For implementation, we choose the ResNet-50 as the backbone and alter the dimensions for the input and output convolutional layers to be consistent with the segmentation task. The network generates one-hot encoded heatmaps and use cross-entropy as the semantic similarity measure.
\section{Experiments}
\label{sec:experiments}
To quantify the benefits that semantic information brings to geometry-based learning, we designed experiments similar to \cite{yin2018geonet}. First, we showed our model's depth prediction performance on KITTI dataset \cite{geiger2012we}, which outperformed state-of-the-art unsupervised and supervised models. Then we designed ablation studies to analyze each individual component's contribution. Finally, we presented improvements in flow predictions and revisited the performance gains using a  category-specific evaluation.

\subsection{Implementation Details}
To make a fair comparison with state-of-the-art models \cite{eigen2014depth, zhou2017unsupervised, yin2018geonet}, we divided KITTI  2015 dataset into train set (40238 images) and test set (697 images) according to the rules from Eigen \textit{et al} \cite{eigen2014depth}. We used DeepLabv3+ \cite{chen2018encoder} (pretrained on \cite{cordts2016cityscapes})  for semantic segmentation and Mask-RCNN \cite{he2017mask} (pretrained on \cite{lin2014microsoft}) for instance segmentation. Similar to the hyper-parameter settings in \cite{yin2018geonet}, we used Adam optimizer \cite{kingma2014adam} with initial learning rate as 2e-4, set batch size to 4 per GPU and trained our modified DepthNet and PoseNet modules for 250000 iterations with random shuffling and data augmentation (random scaling, cropping and RGB perturbation). The training took 10 hours on two GTX1080Ti.

\subsection{Monocular Depth Evaluation on KITTI}
\label{sec:depth_exps}
We augmented the image sequences with corresponding semantic and instance segmentation sequences and adopted the scale normalization suggested in \cite{wang2018learning}. In the evaluation stage, the ground truth depth maps were generated by projecting 3D Velodyne LiDAR points to the image plane. Followed by \cite{yin2018geonet}, we clipped our depth predictions within 0.001m to 80m and calibrated the scale by the medium number of the ground truth. The evaluation results are shown in Table \ref{tab:depth_pred}, where all the metrics are introduced in \cite{eigen2014depth}. Our model benefits significantly from feature augmentation and surpasses the state-of-the-art methods substantially in both supervised and unsupervised fields. 

\begin{figure}[!htbp]
    \centering
    \includegraphics[width=\linewidth]{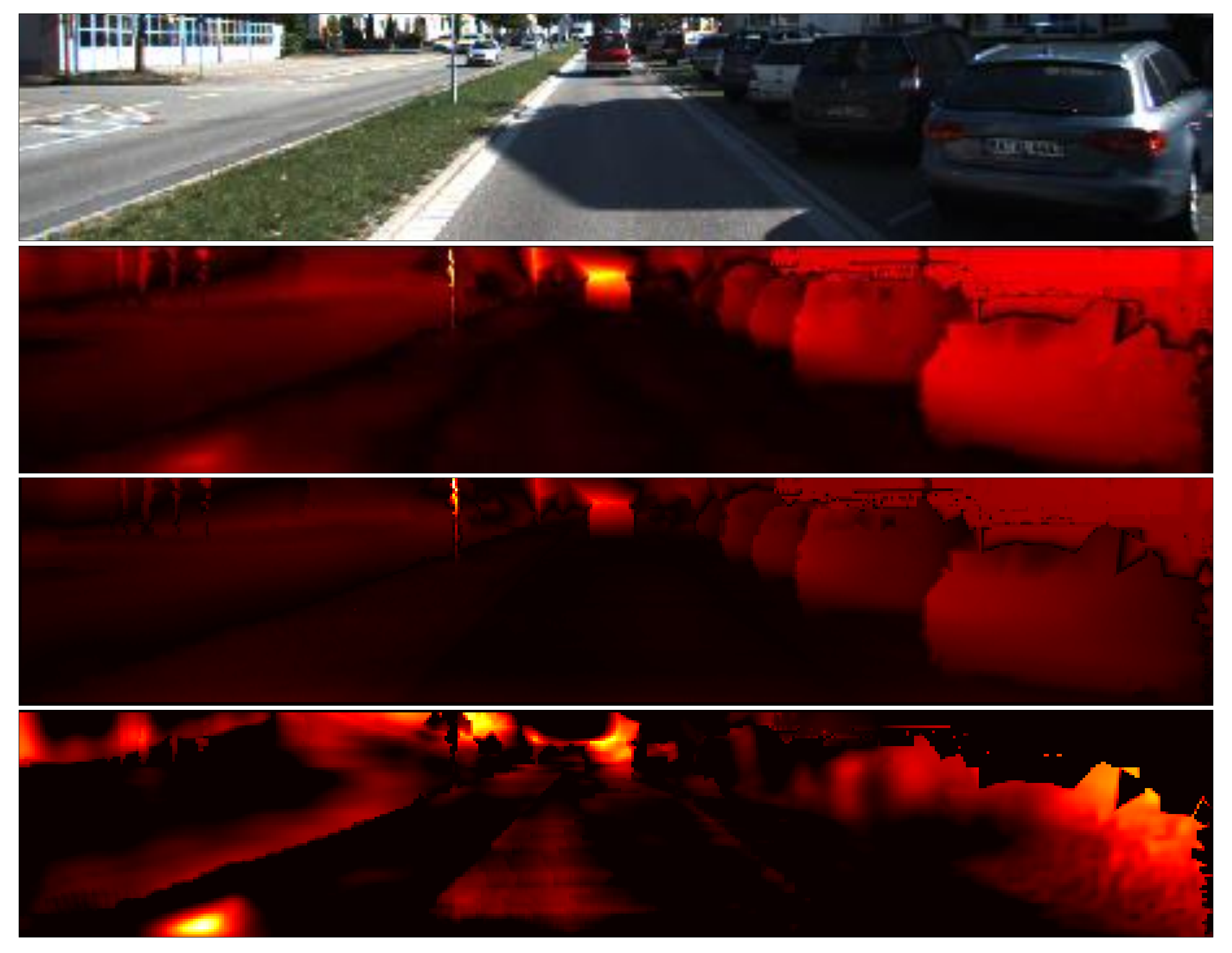}
    \caption{Comparisons of depth evaluations on KITTI. Top to bottom: Input RGB image, AbsRel error map of \cite{yin2018geonet}, AbsRel error map of ours, and improvements of ours on AbsRel map compared to \cite{yin2018geonet}. The ground truth is interpolated to enhance visualization. Lighter color in those heatmaps corresponds to larger errors or improvements.}
    \label{fig:absrel_map}
\end{figure}

Moreover, we found a correlation between the improvement region and object classes. We visualized the absolute relative error (AbsRel) among image plane from our model and from the baseline. As shown in Fig \ref{fig:absrel_map}, most of the improvements come from regions containing objects. This indicates that the network is able to learn the concept of objects to improve the depth prediction by rendering extra semantic information. 

\begin{table*}[htbp]
\centering
\begin{tabular}{ c||c| c c c c|c c c  }
 \hline
  \multirow{2}{*}{Method} & \multirow{2}{*}{Supervised} & \multicolumn{4}{c|}{Error-related metrics} & \multicolumn{3}{c}{Accuracy-related metrics}\\
  &   & Abs Rel & Sq Rel & RSME & RSME log & $\delta < 1.25$ & $\delta < 1.25^2$ & $\delta < 1.25^3$\\
 \hline
 Eigen \textit{et al.} \cite{eigen2014depth} Coarse & Depth & 0.214 & 1.605 & 6.653 & 0.292 & 0.673 & 0.884 & 0.957 \\

  Eigen \textit{et al.} \cite{eigen2014depth} Fine & Depth & 0.203 & 1.548 & 6.307 & 0.282 & 0.702 & 0.890 & 0.957 \\

  Liu \textit{et al.}  \cite{liu2016learning} & Depth & 0.202 & 1.614 & 6.523 & 0.275 & 0.678 & 0.895 & 0.965 \\

  Godard \textit{et al.} \cite{godard2017unsupervised}  & Pose & 0.148 & 1.344 & 5.927 & 0.247 & 0.803 & 0.922 & 0.964 \\

  Zhou \textit{et al.} \cite{zhou2017unsupervised} updated & No & 0.183 & 1.595 & 6.709 & 0.270 & 0.734 & 0.902 & 0.959\\

  Yin \textit{et al.} \cite{yin2018geonet} & No & 0.155 & 1.296 & 5.857 & 0.233 & 0.793 & 0.931 & 0.973 \\

  \hline
  \textbf{Ours} & \multirow{2}{*}{No} & \textbf{0.133} &     \textbf{0.905} &     \textbf{5.181} &     \textbf{0.208} &     \textbf{0.825} &     \textbf{0.947} &     \textbf{0.981}\\
      \textbf{(improved by)}&   & \textbf{14.04\%} &     \textbf{30.19\%} &     \textbf{11.55\%} &    \textbf{10.85\%}  &     \textbf{3.14\%} &     \textbf{1.53\%} &     \textbf{0.80\%} \\
 \hline
\end{tabular}
\caption{Monocular depth results on KITTI 2015 \cite{menze2015object} by the split of Eigen
\textit{et al.} \cite{eigen2014depth} (Our model used scale normalization.)}\label{tab:depth_pred}
\end{table*}

\subsection{Ablation Studies}
\label{sec:ablation}
Here we took a deeper look of our model, testified its robustness under noise from observations, and presented variations of our framework to show promising explorations for future researchers. In the following experiments, we kept all the other parameters the same in \cite{yin2018geonet} and applied the same training/evaluation strategies mentioned in Section \ref{sec:depth_exps}

\parquest{How much gain from various feature augmentation}We tried out different combinations and forms of semantic/instance-level inputs based on ``Yin\textit{ et al}" \cite{yin2018geonet} with scale normalization. From Table \ref{tab:source}, our first conclusion is that any meaningful form of extra input can ameliorate the model, which is straightforward. Secondly, when we use ``Semantic" and ``Instance class" for feature augmentation, one-hot encoding tends to outperform the dense map form. Conceivably one-hot encoding stores richer information in its structural formation, whereas dense map only contains discrete labels which may be more difficult for learning. Moreover, using both ``Semantic" and ``Instance class" can provide further gain, possibly due to the different label distributions of the two datasets. Labels from Cityscape \cite{cordts2016cityscapes} cover both background and foreground concepts, while the COCO dataset \cite{lin2014microsoft} focuses more on objects. At last, when we combined one-hot encoded ``Semantic" and ``Instance class" along with ``Instance id" edge features, the network exploited the most from scene understanding, hence greatly enhanced the performance.

\begin{table*}[htbp]
\centering
\begin{tabular}{ c c c|| c c c c|c c c  }
 \hline
  \multirow{2}{*}{Semantic} &  Instance & Instance & \multicolumn{4}{c|}{Error-related metrics} & \multicolumn{3}{c}{Accuracy-related metrics}\\
  
   &  class & id  & Abs Rel & Sq Rel & RSME & RSME log & $\delta < 1.25$ & $\delta < 1.25^2$ & $\delta < 1.25^3$\\
   
 \hline
  &  & & 0.149 &    1.060 &    5.567    & 0.226 & 0.796 & 0.935 & 0.975\\
  Dense &  &  & 0.142 & 0.991 & 5.309 & 0.216 & 0.814 & 0.943 & 0.980\\

  One-hot &  &  & 0.139& 0.949 & 5.227 & 0.214 & 0.818 & 0.945 & 0.980 \\

   & Dense &  & 0.142 & 0.986 & 5.325 & 0.218 & 0.812 & 0.943 & 0.978 \\

   & One-hot &  & 0.141 & 0.976 & 5.272 & 0.215 & 0.811 & 0.942 & 0.979 \\

   &  & Edge & 0.145 & 1.037 & 5.314 & 0.217 & 0.807 & 0.943 & 0.978 \\
   & Dense & Edge & 0.142 & 0.969 & 5.447 & 0.219 & 0.808 & 0.941 & 0.978 \\
  One-hot & One-hot & Edge &  \textbf{0.133} &     \textbf{0.905} &     \textbf{5.181} &     \textbf{0.208} &     \textbf{0.825} &     \textbf{0.947} &     \textbf{0.981} \\
 \hline
\end{tabular}
\caption{Depth prediction performance gains due to different semantic sources and forms. (Scale normalization was used.)}\label{tab:source}
\end{table*}

\begin{figure}[!htbp]
    \centering
         \begin{subfigure}[b]{1\linewidth}
       \includegraphics[width=\linewidth]{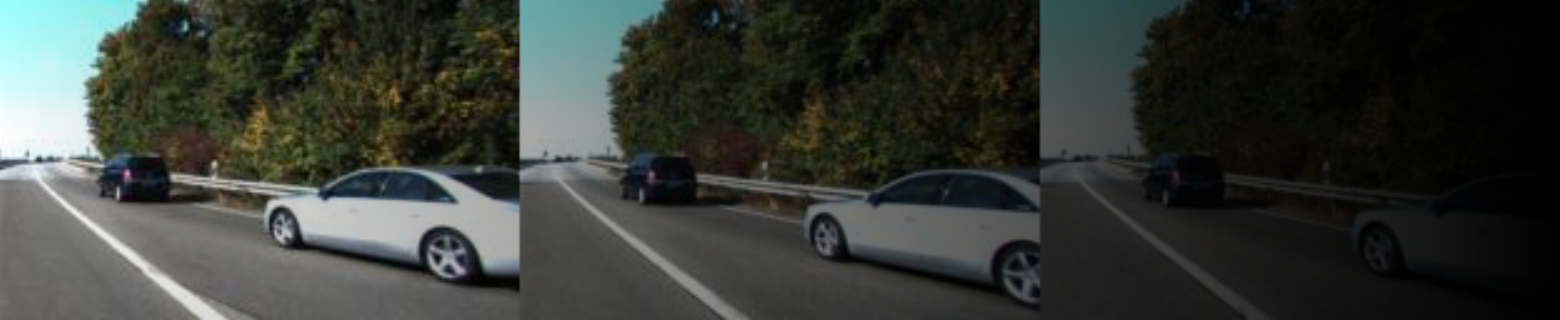}
       \caption{Observations under decreased light condition (left to right)}
       \label{subfig-4:rgb-trans}
     \end{subfigure}
     \begin{subfigure}[b]{1\linewidth}
       \includegraphics[width=\linewidth]{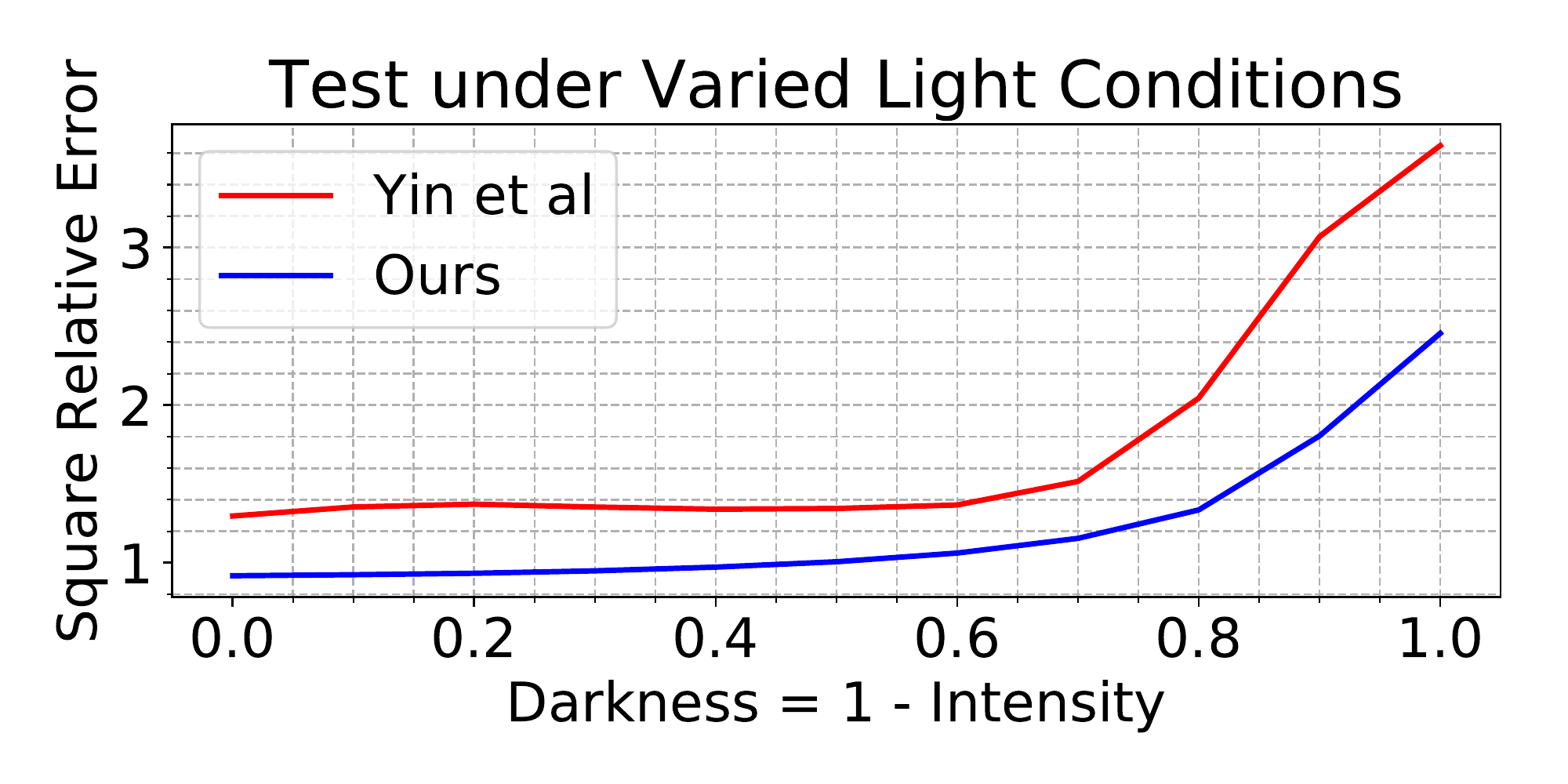}
       \caption{Robustness under decreased light condition}
       \label{subfig-4:noise-test}
     \end{subfigure}
    \caption{The abs errs change as lighting condition drops. Our model can still be better than baseline even if the lighting intensity drops to 0.30 of the original ones.}
    \label{fig:darkness}
    %\vspace{-1em}
\end{figure}

\parquest{Can our model survive under low lighting conditions}
To testify our model's robustness for varied lighting conditions, we multiplied a scalar between 0 and 1 to RGB inputs in the evaluation. Fig \ref{fig:darkness} showed that our model still holds equal performance to \cite{yin2018geonet} when the intensity drops to $30$\%.

\parquest{Which module needs extra information the most} We fed semantics to only DepthNet or PoseNet to see the difference in their performance gain. From Table \ref{tab:module} we can see that compared to DepthNet, PoseNet learns little from the semantics to help depth prediction. Therefore we tried to feed the semantics to a new PoseNet with the same structure as the original one and compute the predicted poses by taking the sum from two different PoseNets, which led to performance gain; however, performance gain was not observed from applying the same method to DepthNet.

\begin{table*}[htbp]
\centering
\begin{tabular}{ c c|| c c c c|c c c  }
 \hline
  \multirow{2}{*}{DepthNet} & 
  \multirow{2}{*}{PoseNet} & \multicolumn{4}{c|}{Error-related metrics} & \multicolumn{3}{c}{Accuracy-related metrics}\\
  
   &  & Abs Rel & Sq Rel & RSME & RSME log & $\delta < 1.25$ & $\delta < 1.25^2$ & $\delta < 1.25^3$\\
   
 \hline
   &  &   0.149 &    1.060 &    5.567    & 0.226 & 0.796 & 0.935 & 0.975 \\
   
  Channel &    & 0.145 & 0.957 & 5.291 & 0.216 & 0.805 & 0.943 & 0.980 \\
   
   &  Channel  & 0.147 & 1.076 & 5.385 & 0.223 & 0.808 & 0.938 & 0.975 \\

  Channel & Channel   & 0.139& 0.949 & \textbf{5.227} & 0.214 & 0.818 & 0.945 & 0.980 \\
  Extra Net & Channel   & 0.147 & 1.036 & 5.593 & 0.226 & 0.803 & 0.937 & 0.975 \\
  Channel & Extra Net & \textbf{0.135} & \textbf{0.932} & 5.241 & \textbf{0.211} & \textbf{0.821} & \textbf{0.945} & \textbf{0.980} \\
 \hline
\end{tabular}
\caption{Each module's contribution toward performance gain from semantics. (Scale normalization was used.)}\label{tab:module}
\end{table*}

\parquest{How to be ``semantic-free" in evaluation}Though semantic helps depth prediction, this idea relies on semantic features during the evaluation phase. If semantic is only utilized in the loss, it would not be needed in evaluation. We attempted to introduce a handcrafted semantic loss term as a weight guidance among image plane but it didn't work well. Also we designed a transfer network which uses the predicted depth to predict semantic maps along with a reconstruction error to help in the training stage. The result in Table \ref{tab:transfer} shows a better result can be obtained by training from pretrained models.

\begin{table*}[htbp]
\centering
\begin{tabular}{ c c|| c c c c|c c c  }
 \hline
  \multirow{2}{*}{Checkpoint} & 
  Transfer & \multicolumn{4}{c|}{Error-related metrics} & \multicolumn{3}{c}{Accuracy-related metrics}\\
  
   & Network & Abs Rel & Sq Rel & RSME & RSME log & $\delta < 1.25$ & $\delta < 1.25^2$ & $\delta < 1.25^3$\\
   
 \hline
  Yin \textit{et al.} \cite{yin2018geonet} &  &   0.155 & 1.296 & 5.857 & 0.233 & 0.793 & 0.931 & 0.973 \\

  Yin \textit{et al.} \cite{yin2018geonet} &  Yes  & 0.150 & 1.141 & 5.709 & 0.231 & 0.792 & 0.934 & 0.974 \\
  
  Yin \textit{et al.} \cite{yin2018geonet} +sn &  &   0.149 &    1.060 &    5.567    & 0.226 & 0.796 & 0.935 & 0.975 \\

  Yin \textit{et al.} \cite{yin2018geonet} +sn &  Yes  & \textbf{0.145} & \textbf{0.994} & \textbf{5.422} & \textbf{0.222} & \textbf{0.806} & \textbf{0.939} & \textbf{0.976} \\
 \hline
\end{tabular}
\caption{Gains in depth prediction using our proposed Transfer Network. (\textbf{+sn}: ``using scale normalization".)}\label{tab:transfer}

%TODO margin
% \vspace*{-18pt}
\end{table*}

\subsection{Optical Flow Estimation on KITTI}
Using our best model for DepthNet and PoseNet in Section \ref{sec:depth_exps}, we conducted rigid flow  and full flow evaluation on KITTI \cite{geiger2012we}. We generated the rigid flow from estimated depth and pose, and compared with \cite{yin2018geonet}. Our model performed better in all the metrics shown in Table \ref{tab:flow}. 
\begin{table}[htbp]
\centering
\begin{tabular}{ c || c c|c c  }
 \hline
    \multirow{2}{*}{Method}&
  \multicolumn{2}{c|}{End Point Error} & \multicolumn{2}{c}{Accuracy}\\
   &  Noc & All & Noc & All\\
 \hline
  Yin \textit{et al.} \cite{yin2018geonet} & 23.5683 & 29.2295 & 0.2345 & 0.2237 \\
  \hline
  Ours   & 22.3819 & 26.8465 & 0.2519 & 0.2376 \\
 \hline
\end{tabular}
\caption{Rigid flow prediction from first stage on KITTI on non-occluded regions(Noc) and overall regions(All).}\label{tab:flow}
\end{table}

\begin{table}[H]%[htbp]
\centering
\begin{tabular}{ c || c c }
 \hline
    \multirow{2}{*}{Method}&
  \multicolumn{2}{c}{End Point Error}\\
   &  Noc & All\\
   \hline
  DirFlowNetS & 6.77 & 12.21 \\
 \hline
  Yin \textit{et al.} \cite{yin2018geonet} & 8.05 & 10.81 \\
  \hline
  Ours   & 7.66 & 13.91 \\
 \hline
\end{tabular}
\caption{Full flow prediction on KITTI 2015 on non-occluded regions(Noc) and overall regions(All). Results from DirFlowNetS are shown in \cite{yin2018geonet}}\label{tab:flow_standard}
\end{table}
We further appended the semantic warping loss introduced in Section \ref{method:semantic_loss} to ResFlowNet in \cite{yin2018geonet} and trained our model on KITTI stereo for 1600000 iterations. As demonstrated in Table \ref{tab:flow_standard}, flow prediction got improved in non-occluded region compared to  \cite{yin2018geonet} and our model produced comparable results in overall regions.

\subsection{Category-Specific Metrics Evaluation}
\begin{figure}
    \centering
         \begin{subfigure}[b]{0.49\linewidth}
       \includegraphics[width=\linewidth]{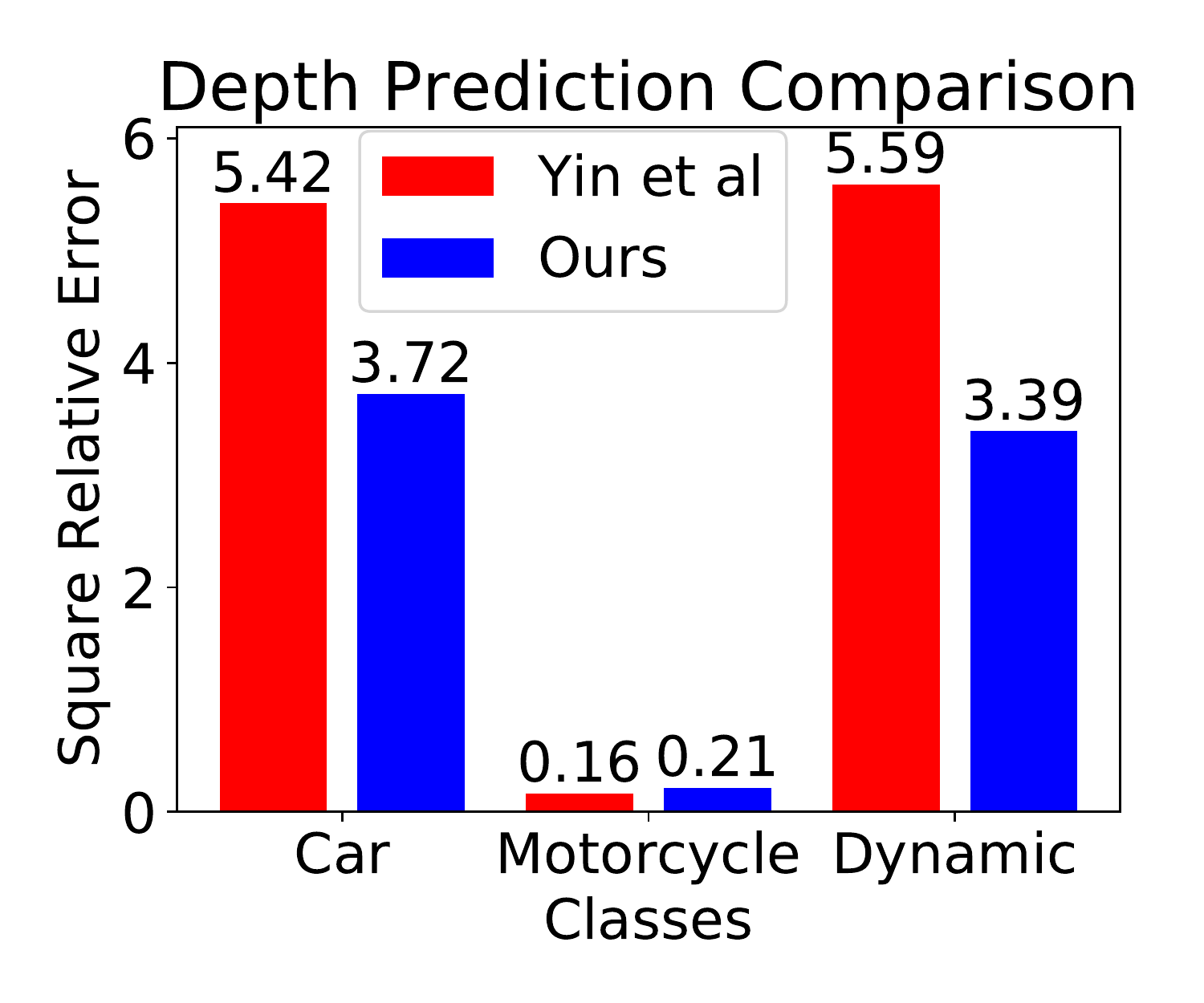}
     \end{subfigure}
     \begin{subfigure}[b]{0.49\linewidth}
       \includegraphics[width=\linewidth]{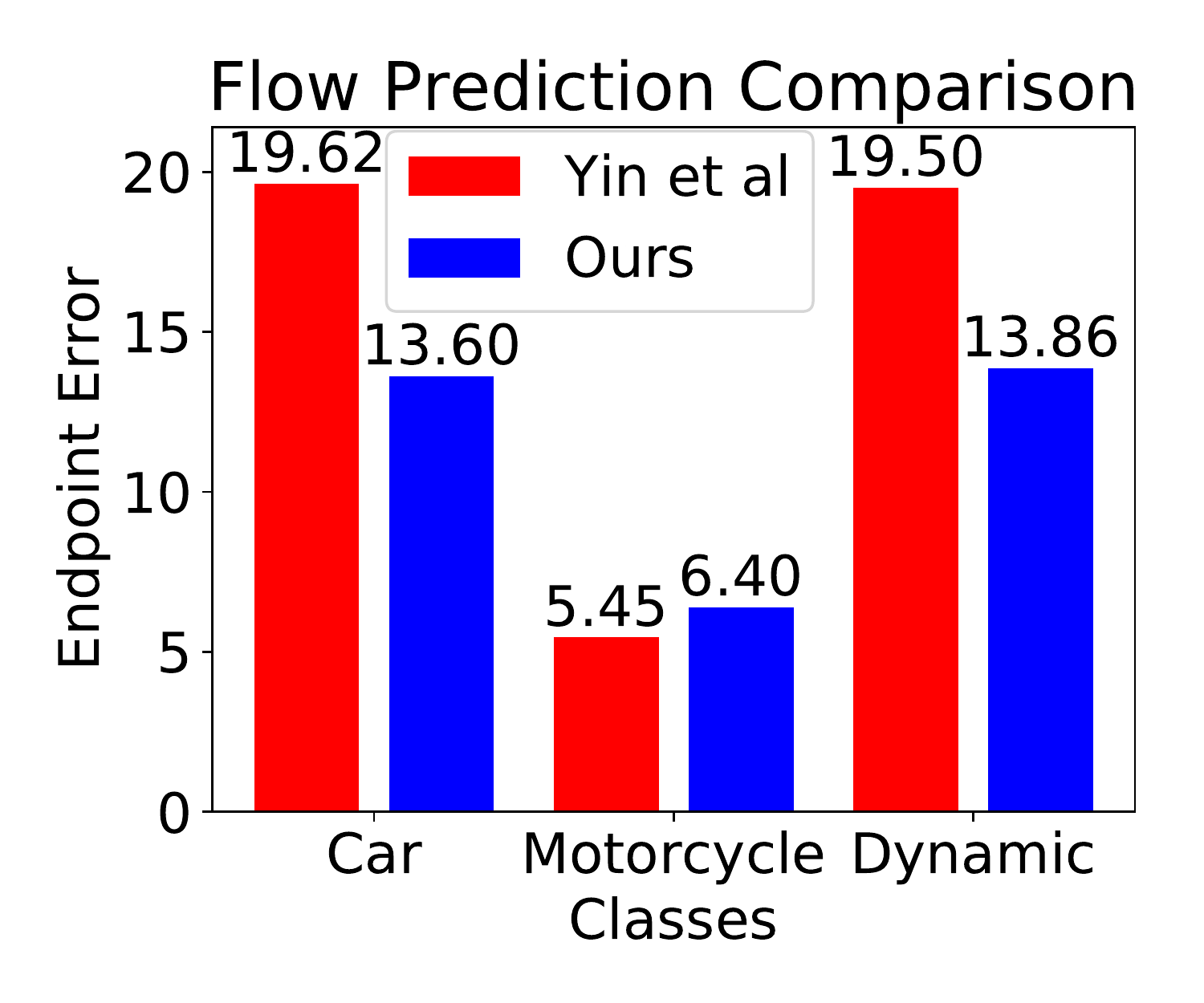}
     \end{subfigure}
    % \vspace{-1.5em}
    \caption{Performance gains in depth (left) and flow (right) among different classes of dynamic objects.}
    \label{fig:class-eval}
    
\end{figure}

This section will present the improvements by semantic categories. As shown in the bar-chart in Fig \ref{fig:class-eval}, most improvements were shown in ``Vehicle" and ``Dynamic" classes\footnote{For ``Dynamic" classes, we choose ``person", ``rider", ``car", ``truck", ``bus", ``train", ``motorcycle" and ``bicycle" classes as defined in \cite{cordts2016cityscapes}}, where errors are generally large. Our network did not improve much for other less frequent categories, such as ``Motorcycle", which are generally more difficult to segment in images.

\section{Conclusion}
\label{sec:conclusion}
In \name, we strive to achieve robust performance for depth and flow perception without using geometric labels. To achieve this goal, \name utilizes semantic and instance segmentation to create spatial constraints on the geometric attributes of the pixels. We present novel methods of feature augmentation and loss augmentation to include semantic labels in the geometry predictions. This work presents a first of a kind approach which moves away from pixel-level to object-level depth and flow predictions. Most notably, our method significantly surpasses the state-of-the-art solution for monocular depth estimation. In the future, we would like to extend our \name to various sensor modalities (IMU, LiDAR or thermal).\\

\partitle{Acknowledgement}This work was supported by UCSD faculty startup (Prof. Bharadia), Toyota InfoTechnology Center and Center for Wireless communication at UCSD.

%%%%%%%%% BIBLIOGRAPHY
\FloatBarrier
\twocolumn
{\small
\bibliographystyle{ieee_fullname}
\bibliography{main}
}

\newpage
\newpage
\onecolumn
\setcounter{section}{0}
\setcounter{figure}{0}
\setcounter{table}{0}
\renewcommand\thesection{\Alph{section}}
\renewcommand\thesubsection{\Alph{subsection}}

\newcommand{\TODO}[1]{\textbf{\textcolor{red}{(: #1)}}}
\newcommand{\CAPOUR}{\noindent Top to bottom: input image, semantic segmentation, instance segmentation, ground truth disparity map, disparity prediction from baseline(Yin \textit{et al}. \cite{yin2018geonet}) , disparity prediction from ours, AbsRel error map of baseline models, AbsRel error map of ours and the improvement region compared to baseline. For the purpose of visualization, disparity maps are interpolated and cropped\cite{Garg2016UnsupervisedCF}. For all heatmaps, darker means smaller value (disparity, error or improvement). Typical image regions where we do better include cars, pedestrians and other common dynamic objects}

\newcommand{\GAP}{\quad}

{\Large\noindent  \textbf{Supplementary Material for \name}\vspace{8pt}}

\noindent \textbf{Additional ablation studies on loss augmentations:} As mentioned in our paper, the heuristic loss functions are not effective even after careful hyper-parameter tuning. This motivated us to design a learnable loss function (transfer network), which does improve upon the baseline as shown in Table \ref{tab:transfer} of our paper.

\begin{table}[!htbp]
\vspace*{3pt}
\centering

\begin{tabular}{c|cccc|ccc}
 \hline
  \multirow{2}{*}{Method} & \multicolumn{4}{c|}{Error metrics} & \multicolumn{3}{c}{Accuracy $(\delta<)$}\\
  
  & AbsRel & SqRel & RSME & RSME\textsubscript{log} & $1.25^1$ & $1.25^2$ & $1.25^3$\\

 \hline
  Yin \textit{et al.}\cite{yin2018geonet}  &   0.155 & 1.296 & 5.857 & 0.233 & \textbf{0.793} & 0.931 & 0.973 \\
  Warp Loss   & 0.169 & 1.246 & 6.233 & 0.254 & 0.750 & 0.917 & 0.968 \\
  Mask Loss   & 0.165 & 1.204 & \textbf{5.593} & 0.232 & 0.769 & 0.926 &  0.974 \\
  Edge Loss   & 0.163 & 1.230 & 5.961 & 0.243 & 0.774 & 0.924 & 0.970 \\ 
  Transfer    & \textbf{0.150} & \textbf{1.141} & 5.709 & \textbf{0.231} & 0.792 & \textbf{0.934} & \textbf{0.974} \\
 \hline
\end{tabular}
\vspace*{1pt}
\caption{Depth predictions for different loss augmentations (without using scale normalization). Here Warp Loss, Mask Loss and Edge Loss are on par or not as good as the baseline, whereas Transfer Network surpasses the baseline in almost all the metrics.}
%TODO margin
%\vspace*{-18pt}
\label{tab:miss_exp}
\end{table}

\begin{figure}[!htbp]
\begin{center}
   \includegraphics[width=0.55\linewidth]{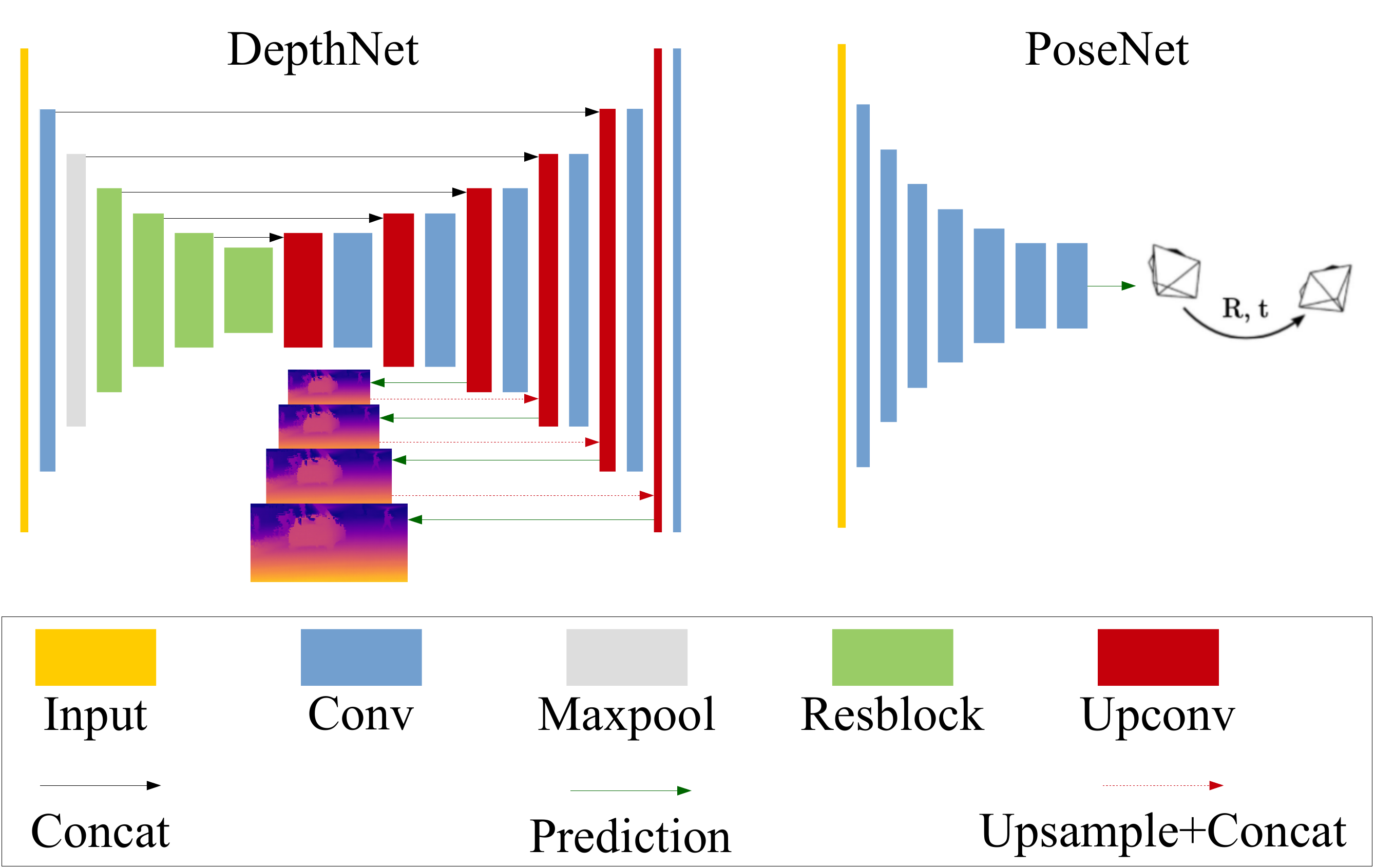}
\end{center}
\vspace*{-12pt}
   \caption{Network structures for DepthNet and PoseNet}
\label{fig:network_comp}
\vspace*{4pt}
\end{figure}

\noindent \textbf{Why does ”ExtraNet” only work for PoseNet?}
In the ablation studies in Section \ref{sec:ablation}, we tested the contribution of semantic information in each module. The result suggests that vanilla PoseNet benefits from  semantics  only  marginally,  which  might  due  to  its simple  structure. By adding Extra  Network (”ExtraNet”) to PoseNet, our model gained further improvement. ”ExtraNet” does not benefit DepthNet because the latter has already had a complicated structure as shown in Figure \ref{fig:network_comp}. \\

\noindent\textbf{More visualization results for depth estimation:} In the rest of the supplementary material, we will present extra visualization results to help readers understand where our semantic-aided model improved the most. We compared the prediction result from our best model in Table \ref{tab:depth_pred} with Yin \textit{et al}. \cite{yin2018geonet} and ground truth. We followed \cite{Garg2016UnsupervisedCF} to plot the prediction result using disparity heatmaps. The following results show that our model can gain improvement from regions belonging to cars and other dynamic classes.

\newpage
\begin{figure}[!hbtp]
    \centering
         \begin{subfigure}[b]{0.48\linewidth}
      \includegraphics[width=\linewidth]{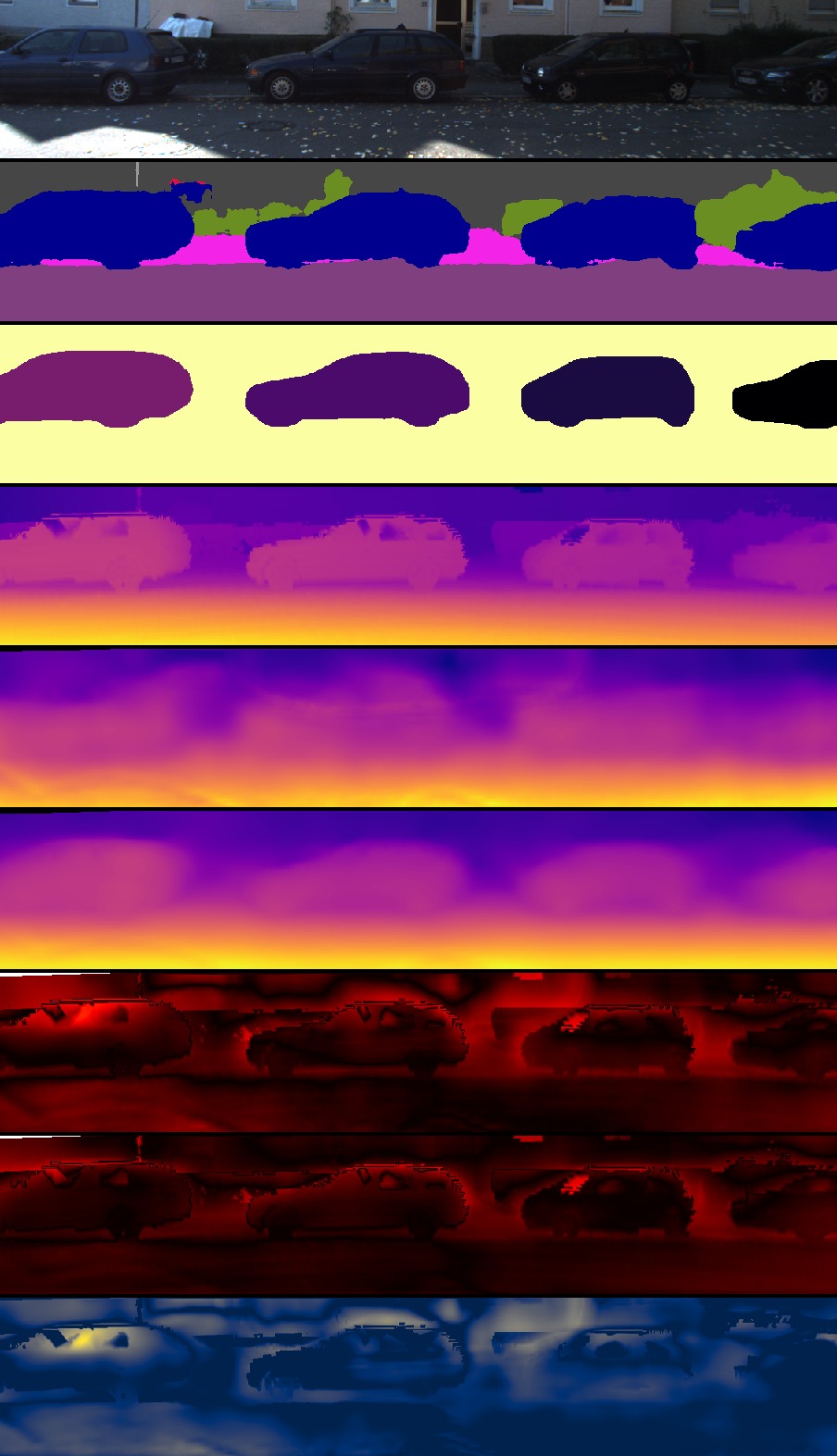}
     \end{subfigure}
     \GAP
     \begin{subfigure}[b]{0.48\linewidth}
      \includegraphics[width=\linewidth]{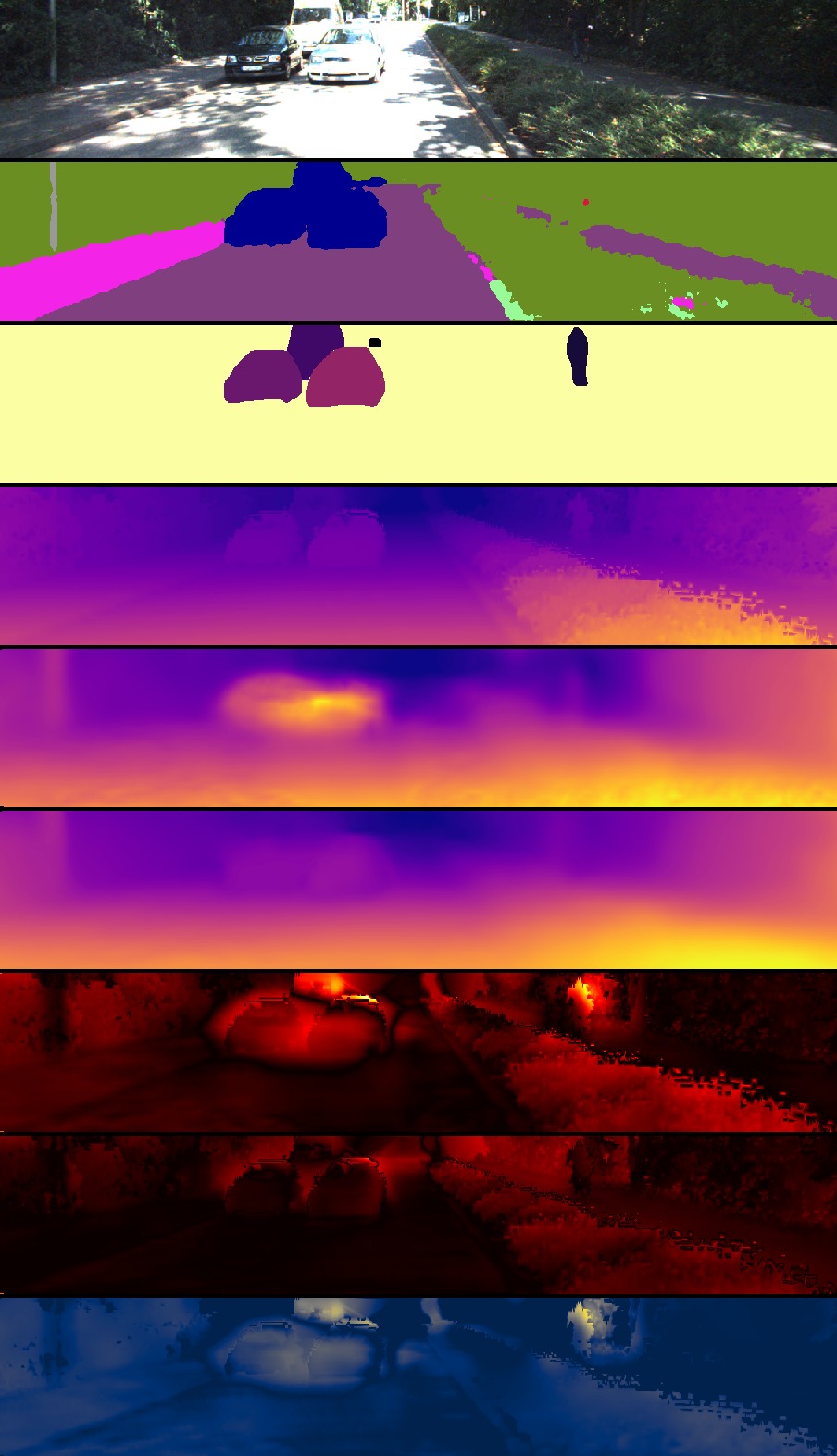}
     \end{subfigure}
    \caption{\CAPOUR}
\end{figure}

\begin{figure}[!hbtp]
    \centering
         \begin{subfigure}[b]{0.48\linewidth}
      \includegraphics[width=\linewidth]{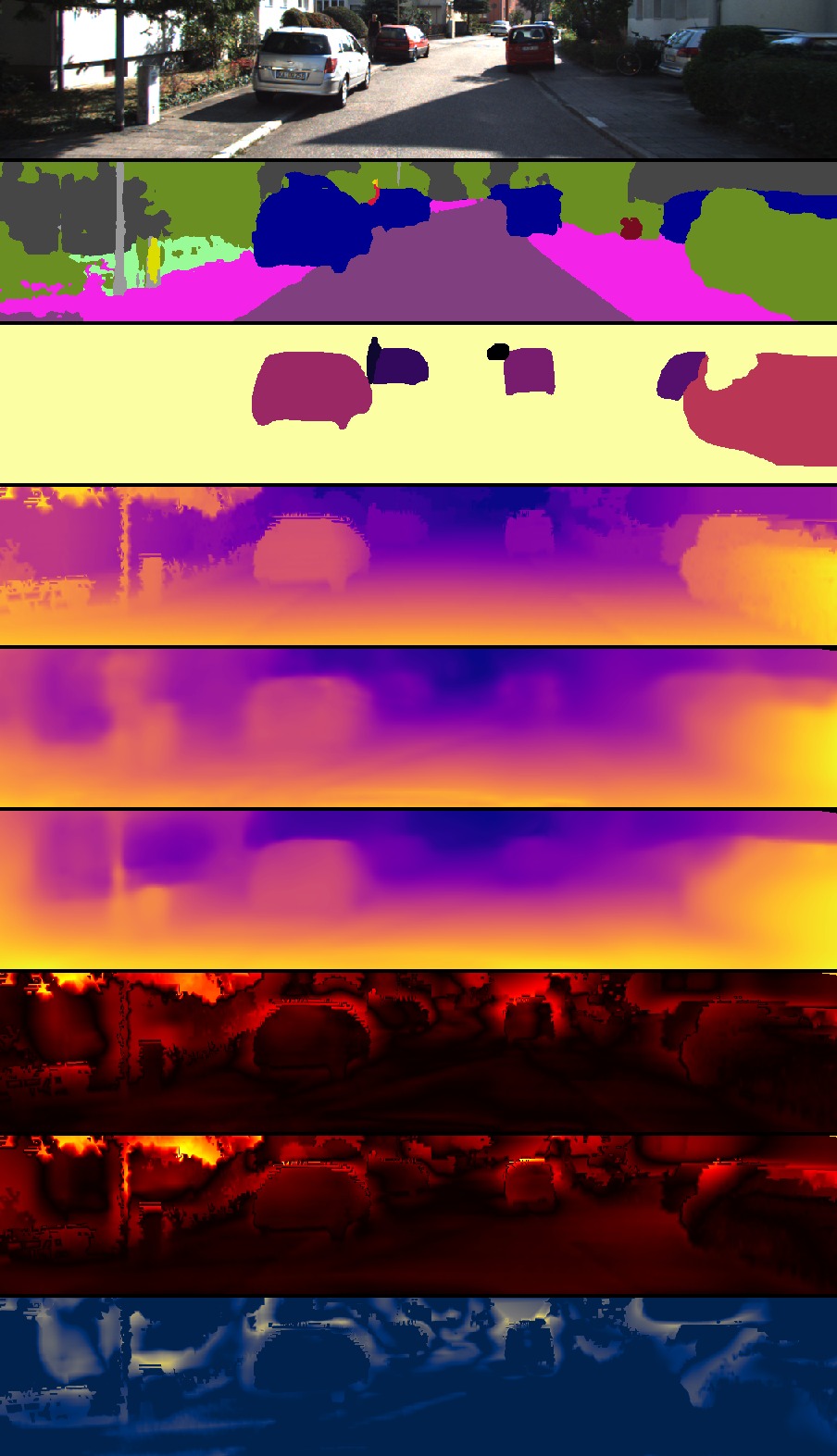}
     \end{subfigure}
     \GAP
     \begin{subfigure}[b]{0.48\linewidth}
      \includegraphics[width=\linewidth]{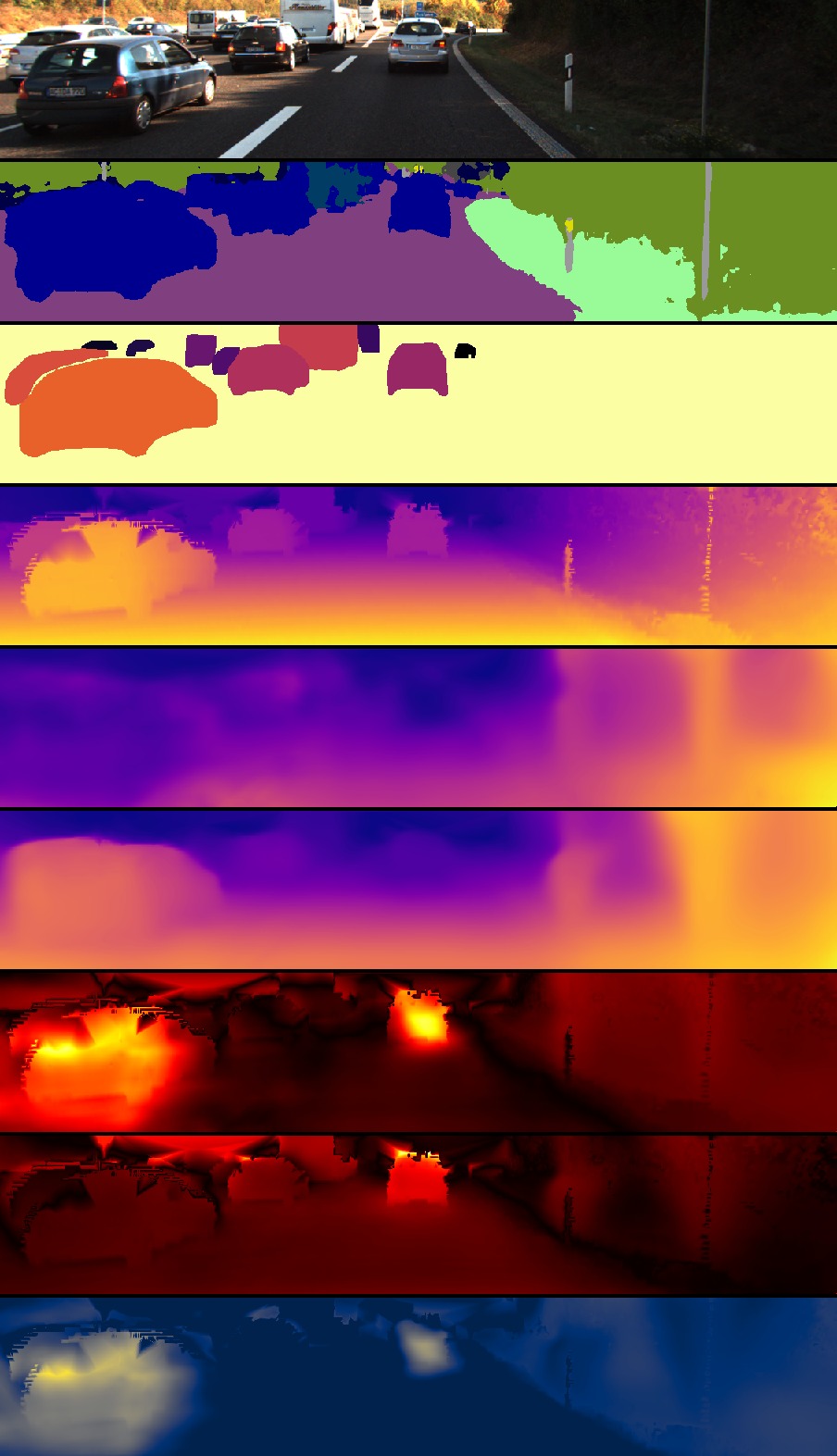}
     \end{subfigure}
    \caption{\CAPOUR}
\end{figure}

\begin{figure}[!hbtp]
    \centering
         \begin{subfigure}[b]{0.48\linewidth}
      \includegraphics[width=\linewidth]{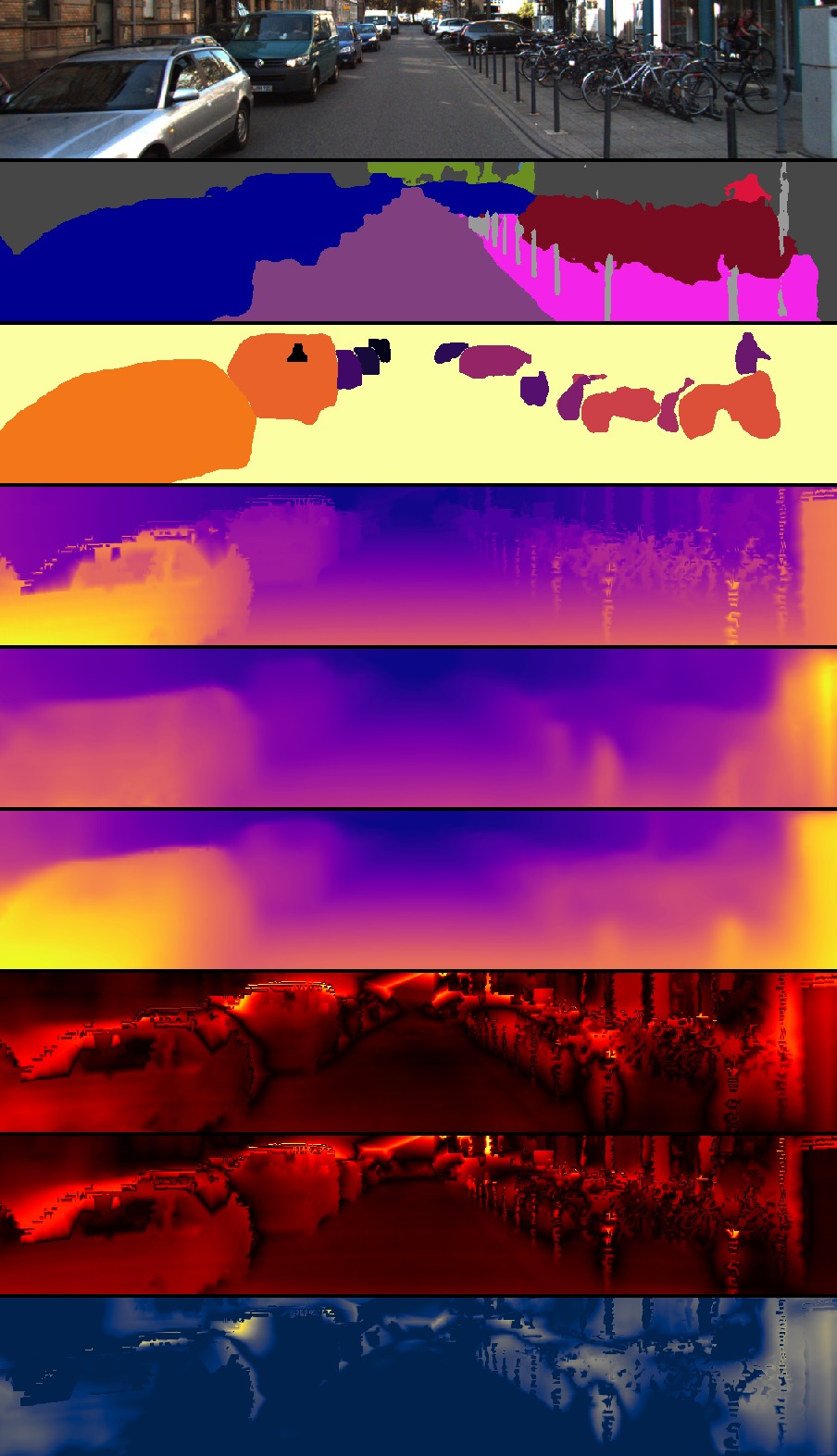}
     \end{subfigure}
     \GAP
     \begin{subfigure}[b]{0.48\linewidth}
      \includegraphics[width=\linewidth]{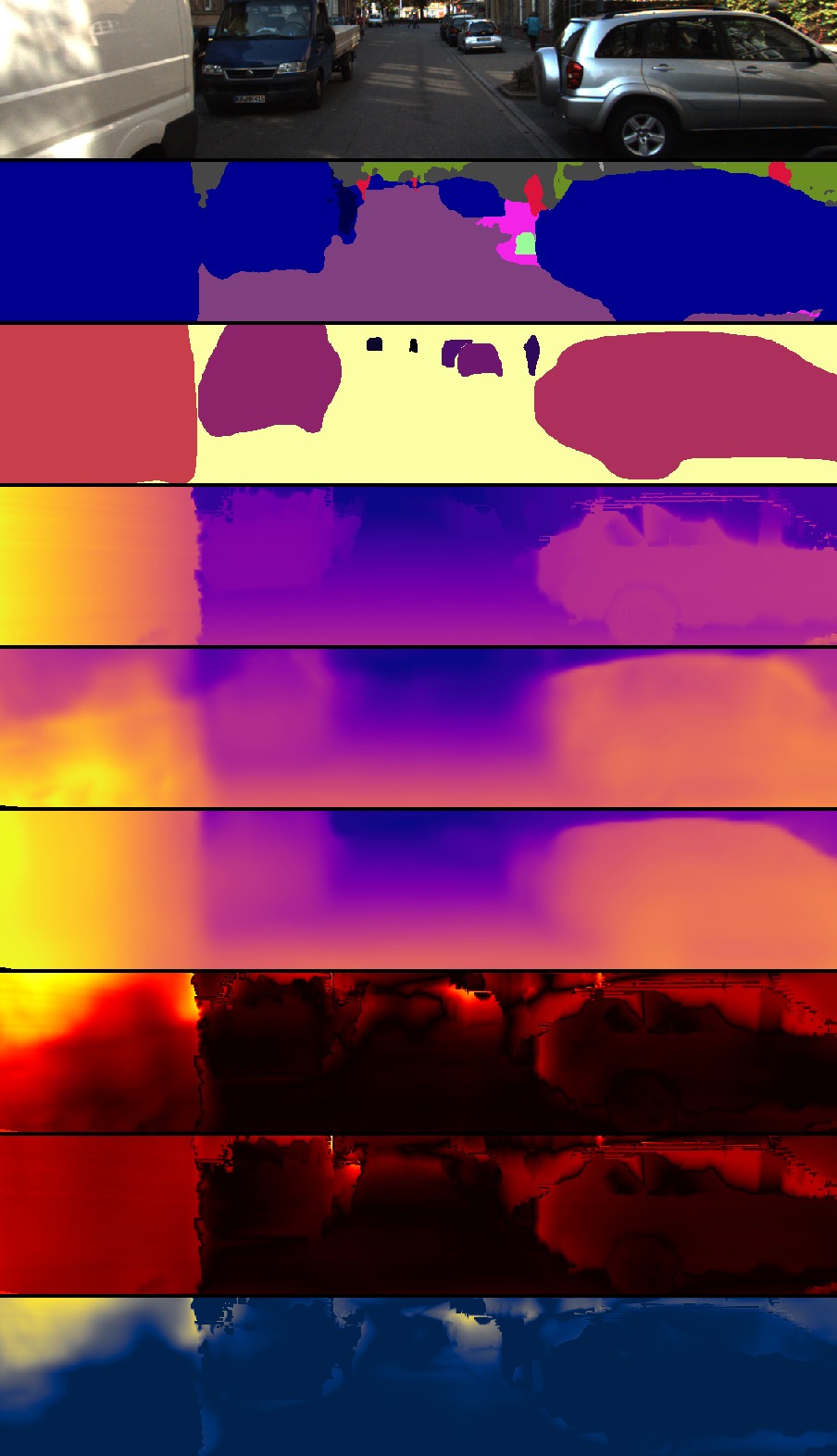}
     \end{subfigure}
    \caption{\CAPOUR}
\end{figure}

\begin{figure}[!hbtp]
    \centering
         \begin{subfigure}[b]{0.48\linewidth}
      \includegraphics[width=\linewidth]{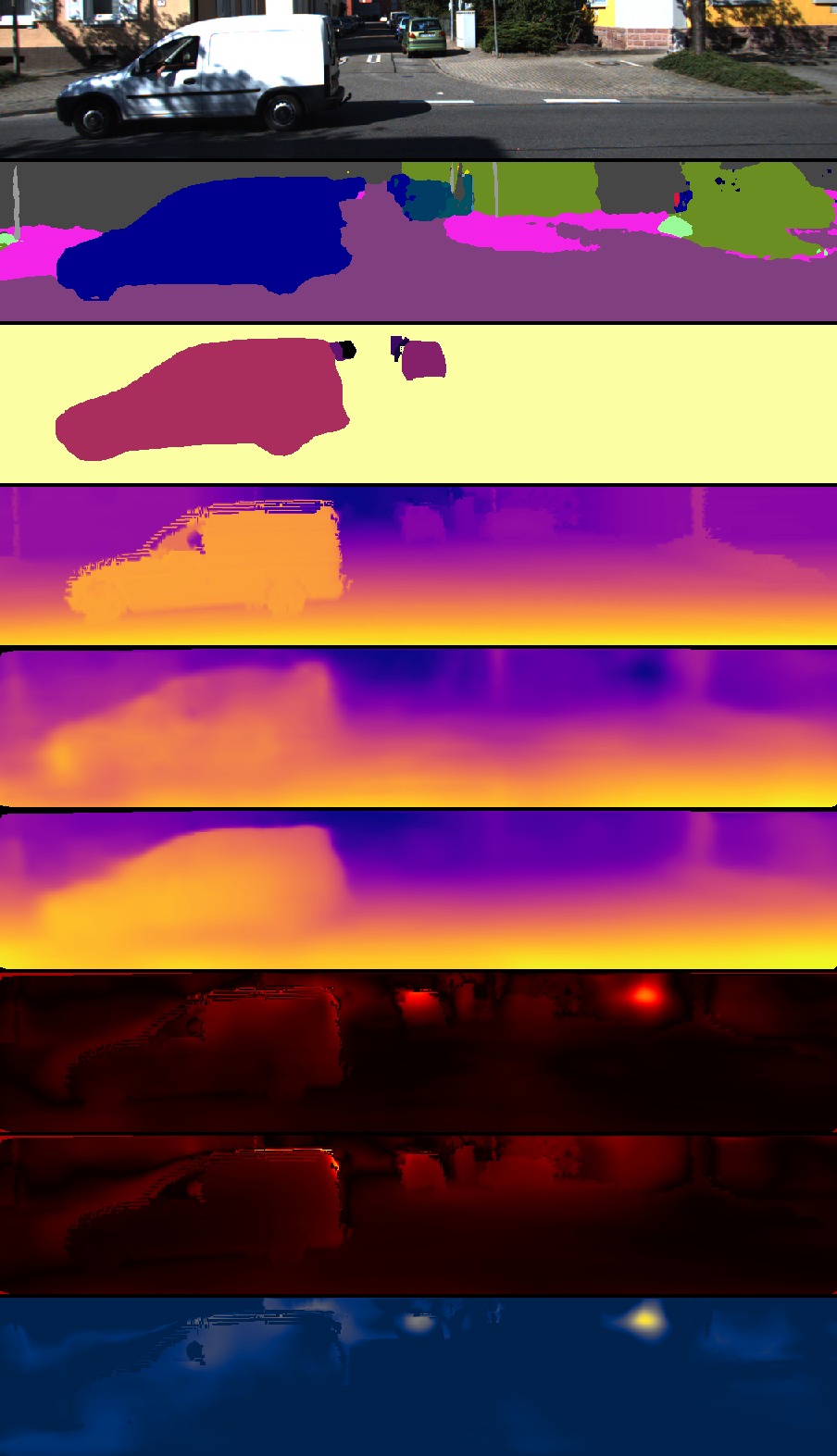}
     \end{subfigure}
     \GAP
     \begin{subfigure}[b]{0.48\linewidth}
      \includegraphics[width=\linewidth]{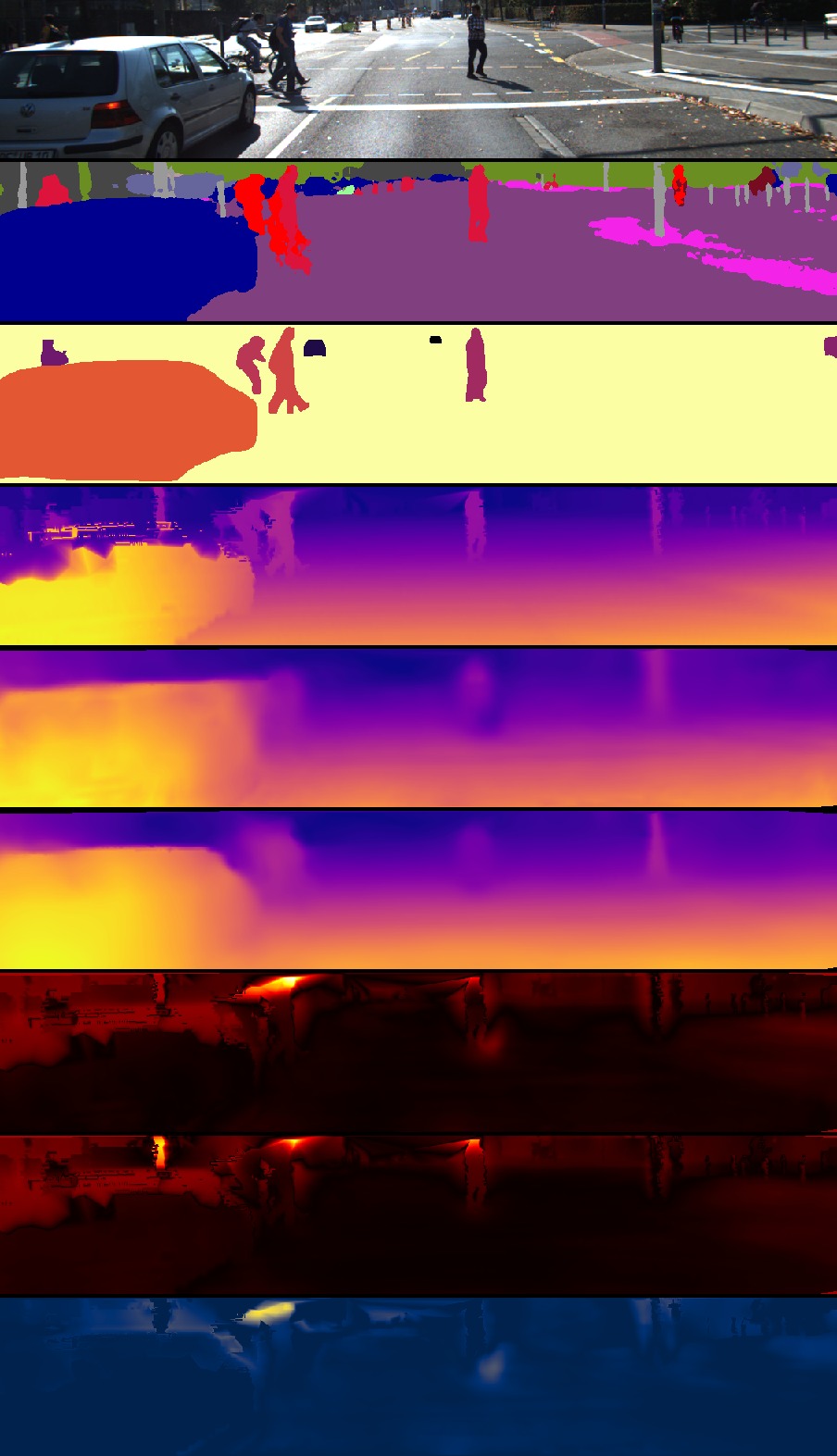}
     \end{subfigure}
    \caption{\CAPOUR}
\end{figure}

\begin{figure}[!hbtp]
    \centering
         \begin{subfigure}[b]{0.48\linewidth}
      \includegraphics[width=\linewidth]{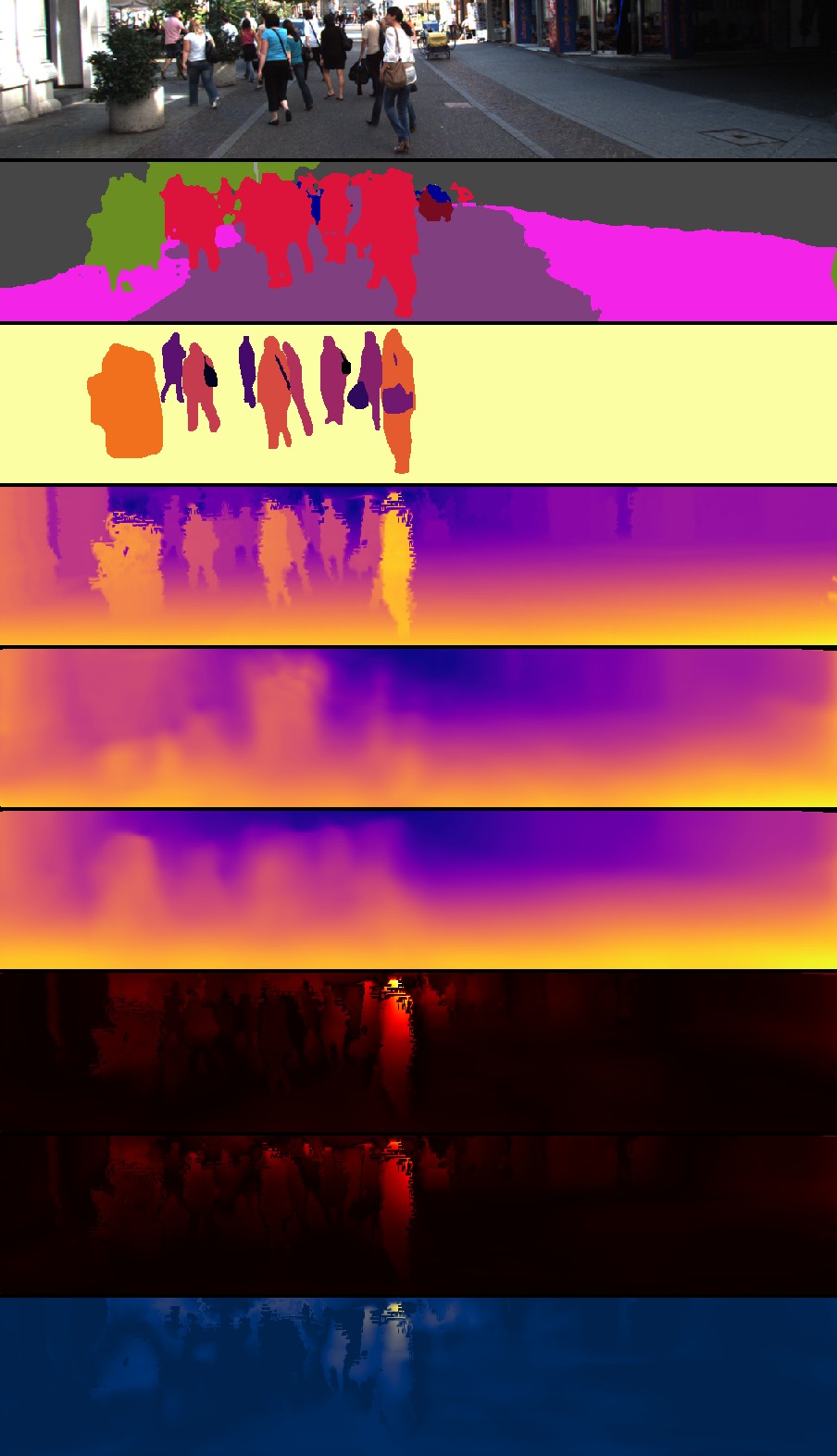}
     \end{subfigure}
     \GAP
     \begin{subfigure}[b]{0.48\linewidth}
      \includegraphics[width=\linewidth]{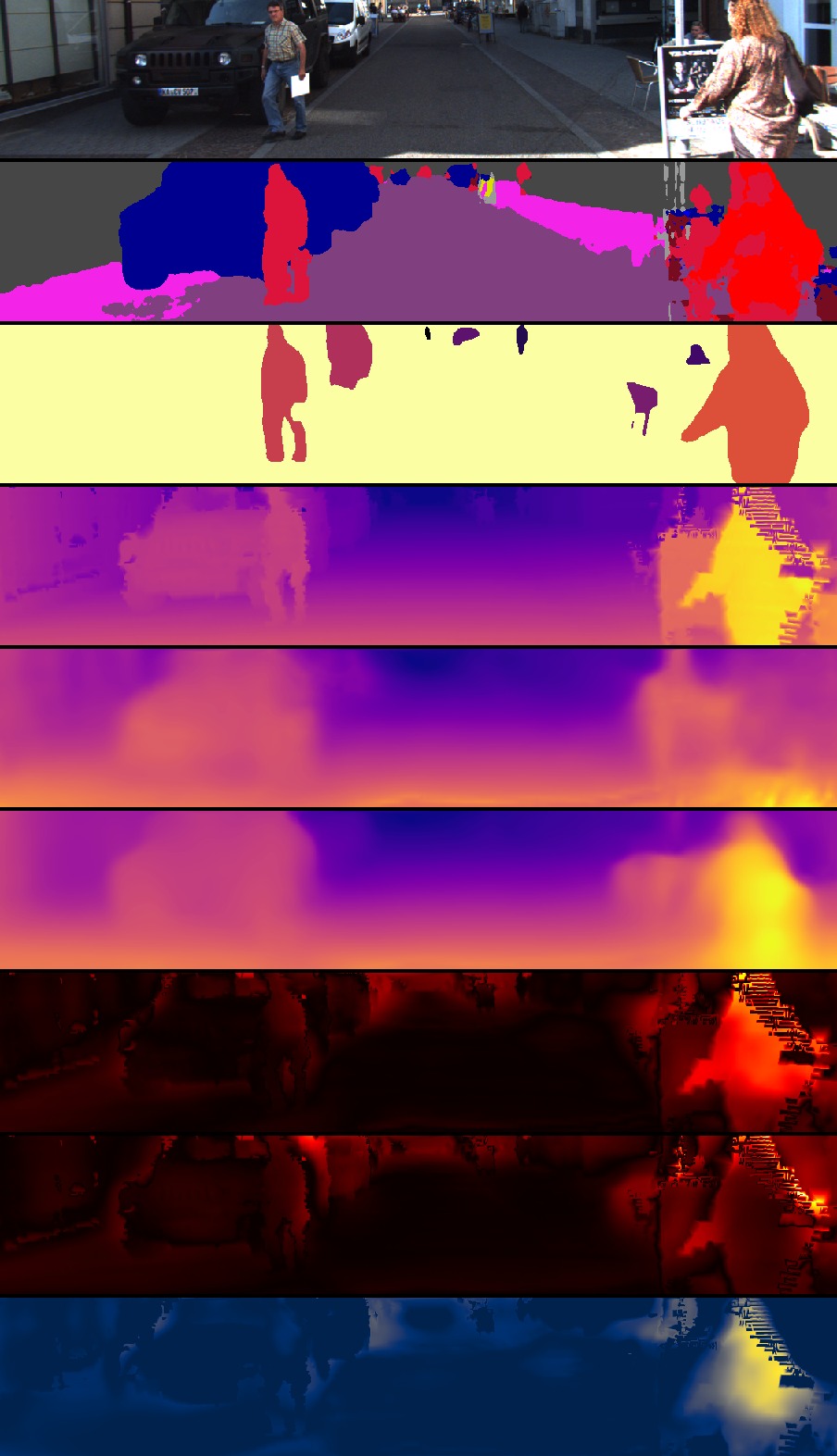}
     \end{subfigure}
    \caption{\CAPOUR}
\end{figure}

\begin{figure}[!hbtp]
    \centering
         \begin{subfigure}[b]{0.48\linewidth}
      \includegraphics[width=\linewidth]{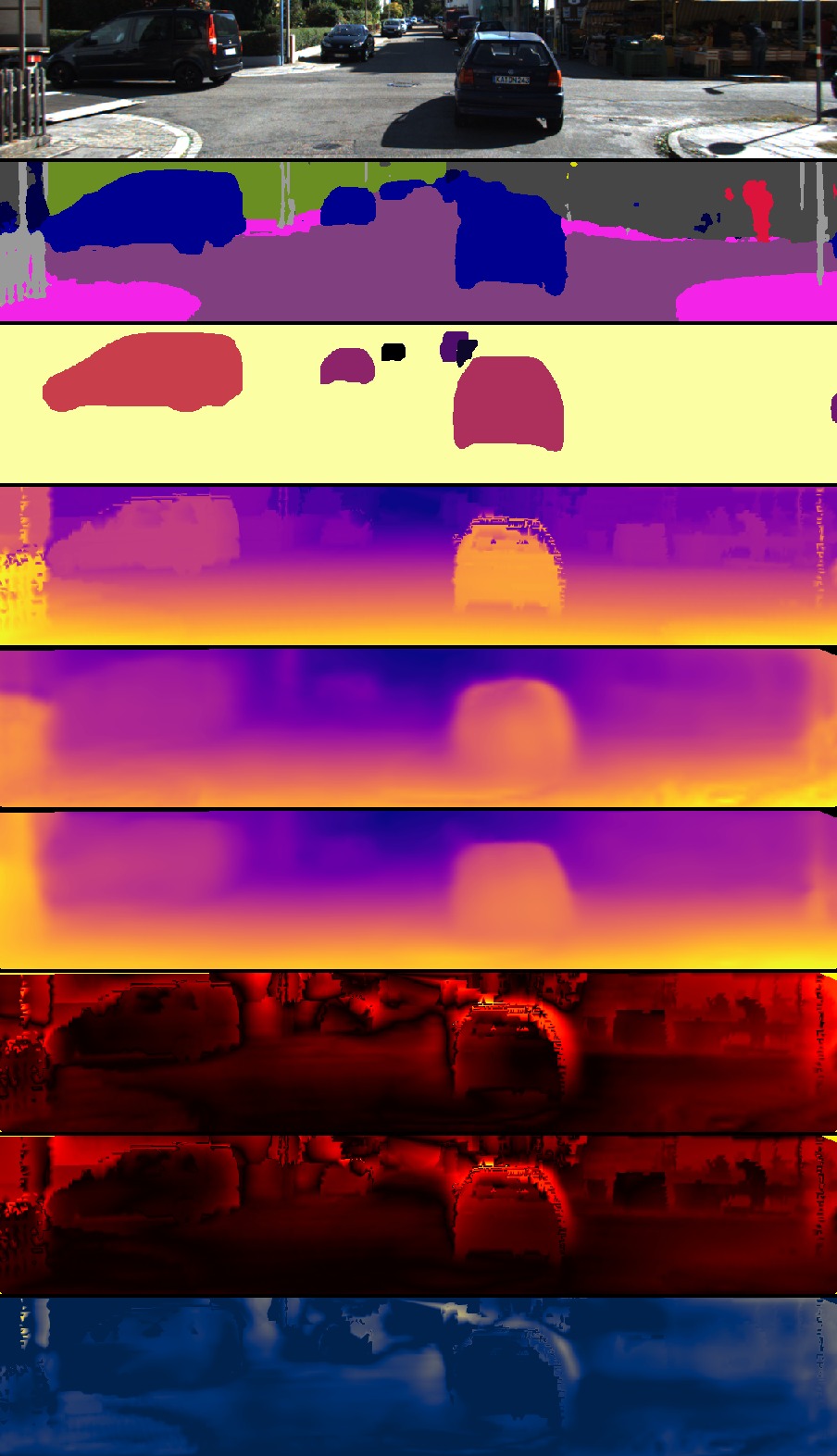}
     \end{subfigure}
     \GAP
     \begin{subfigure}[b]{0.48\linewidth}
      \includegraphics[width=\linewidth]{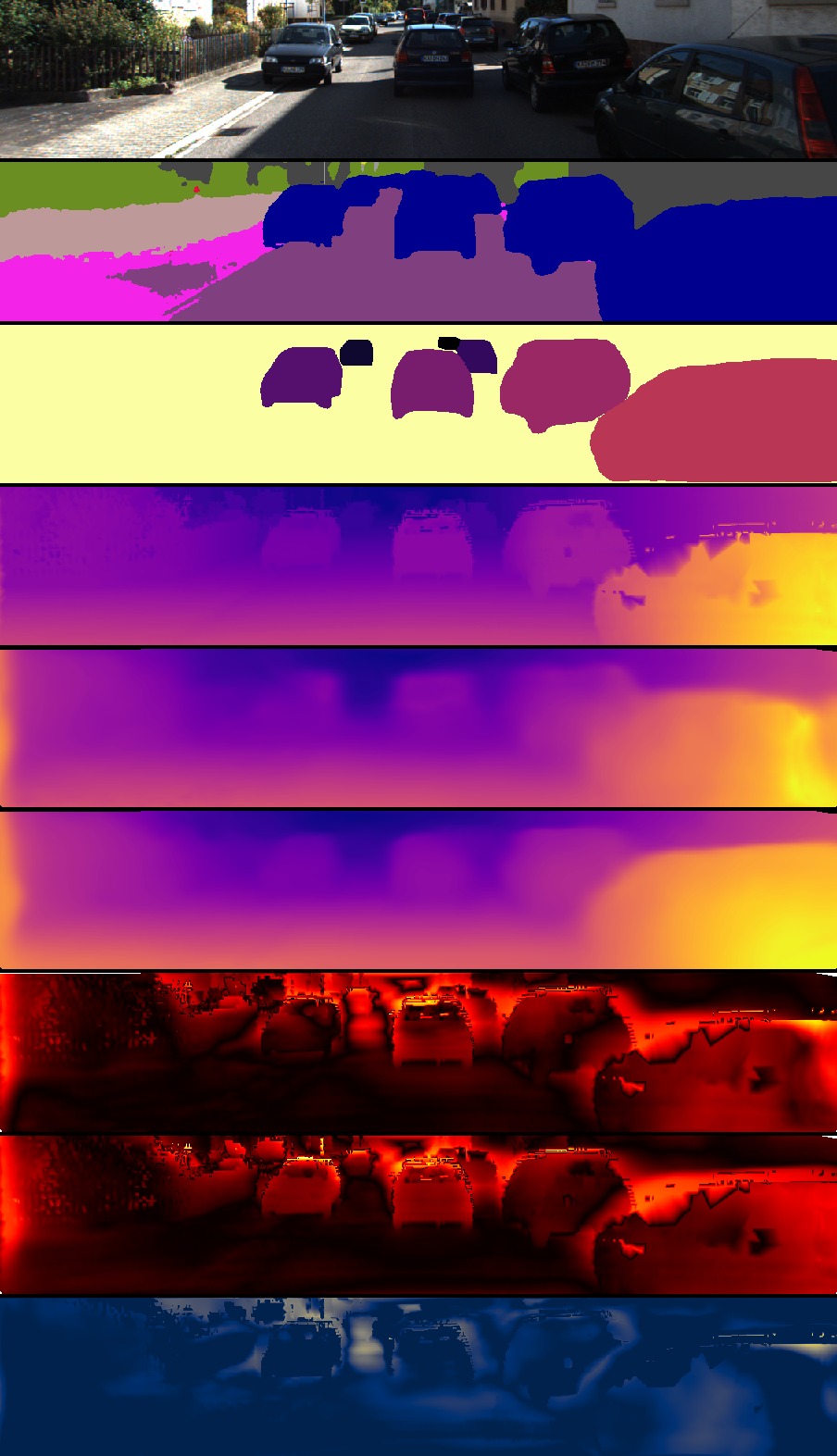}
     \end{subfigure}
    \caption{\CAPOUR}
\end{figure}

\begin{figure}[!hbtp]
    \centering
         \begin{subfigure}[b]{0.48\linewidth}
      \includegraphics[width=\linewidth]{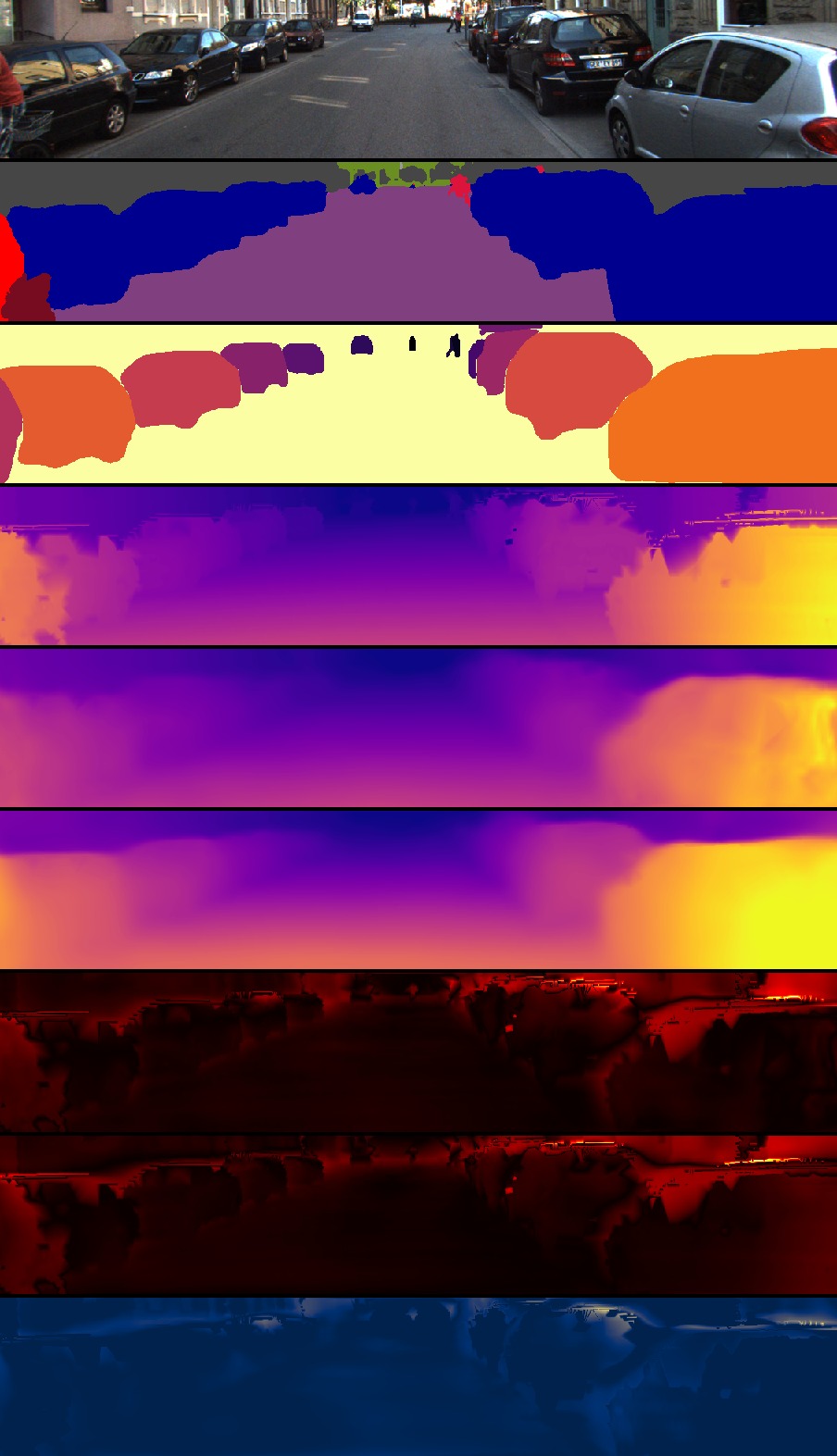}
     \end{subfigure}
     \GAP
     \begin{subfigure}[b]{0.48\linewidth}
      \includegraphics[width=\linewidth]{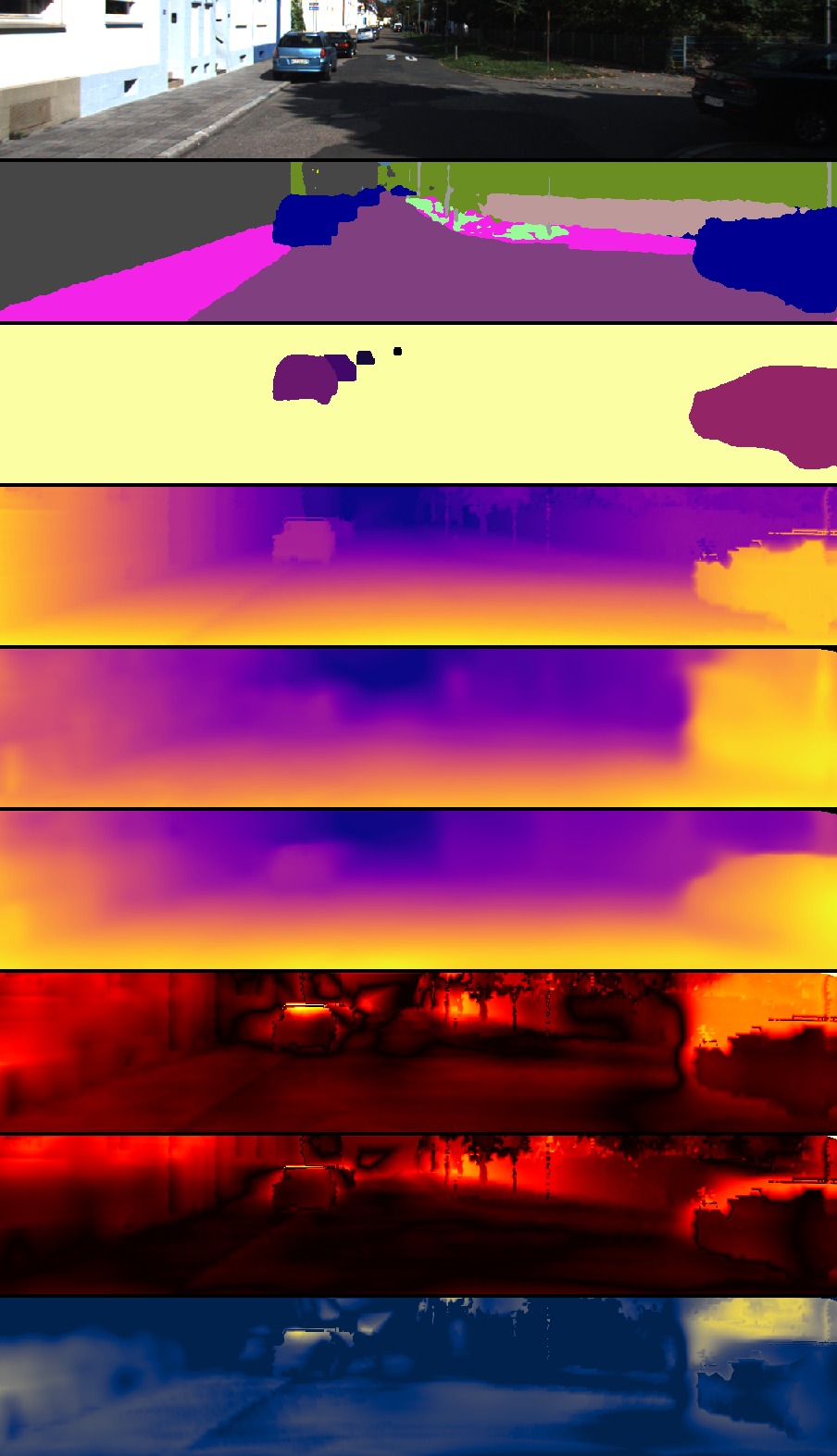}
     \end{subfigure}
    \caption{\CAPOUR}
\end{figure}

\begin{figure}[!hbtp]
    \centering
         \begin{subfigure}[b]{0.48\linewidth}
      \includegraphics[width=\linewidth]{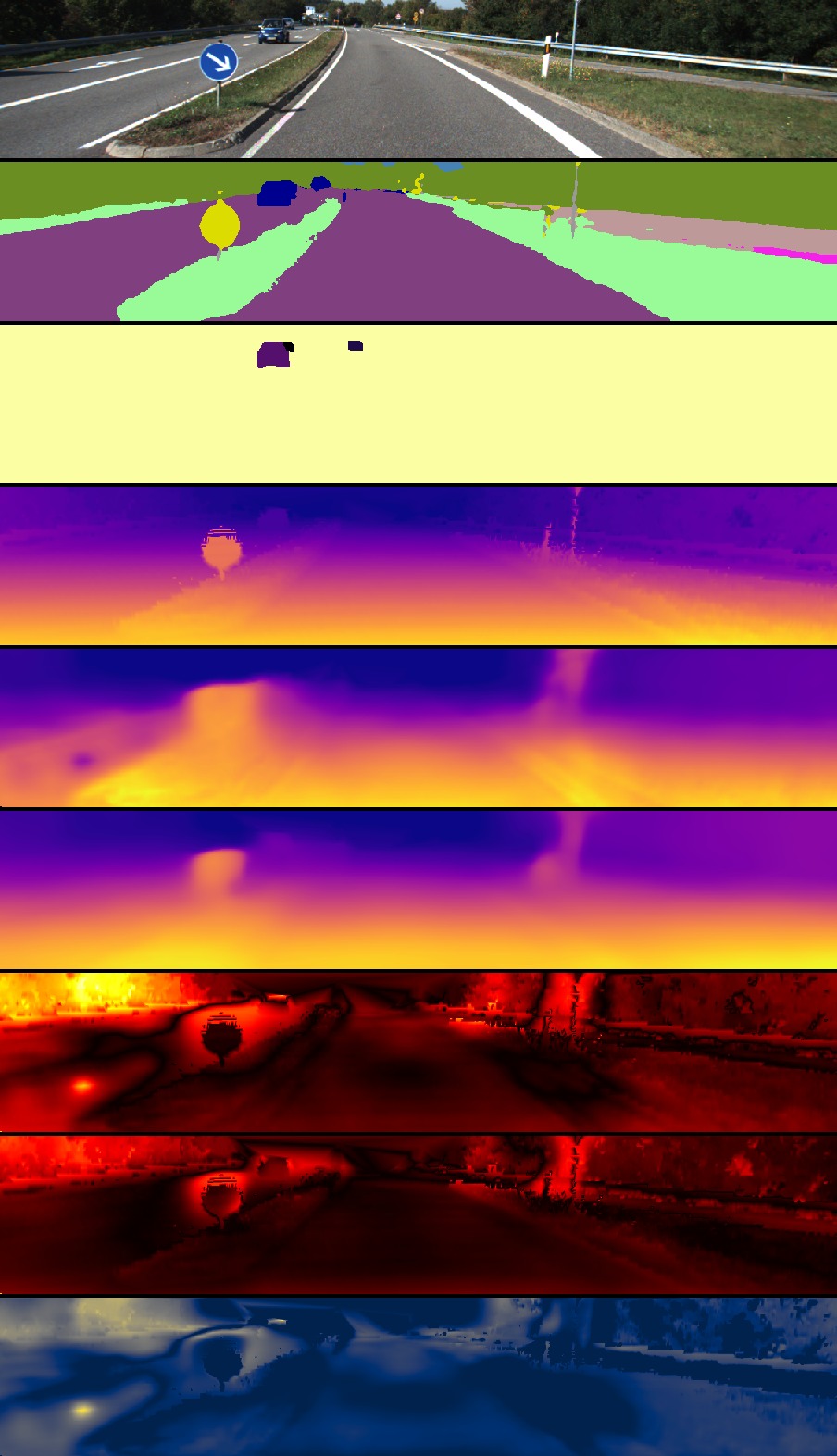}
     \end{subfigure}
     \GAP
     \begin{subfigure}[b]{0.48\linewidth}
      \includegraphics[width=\linewidth]{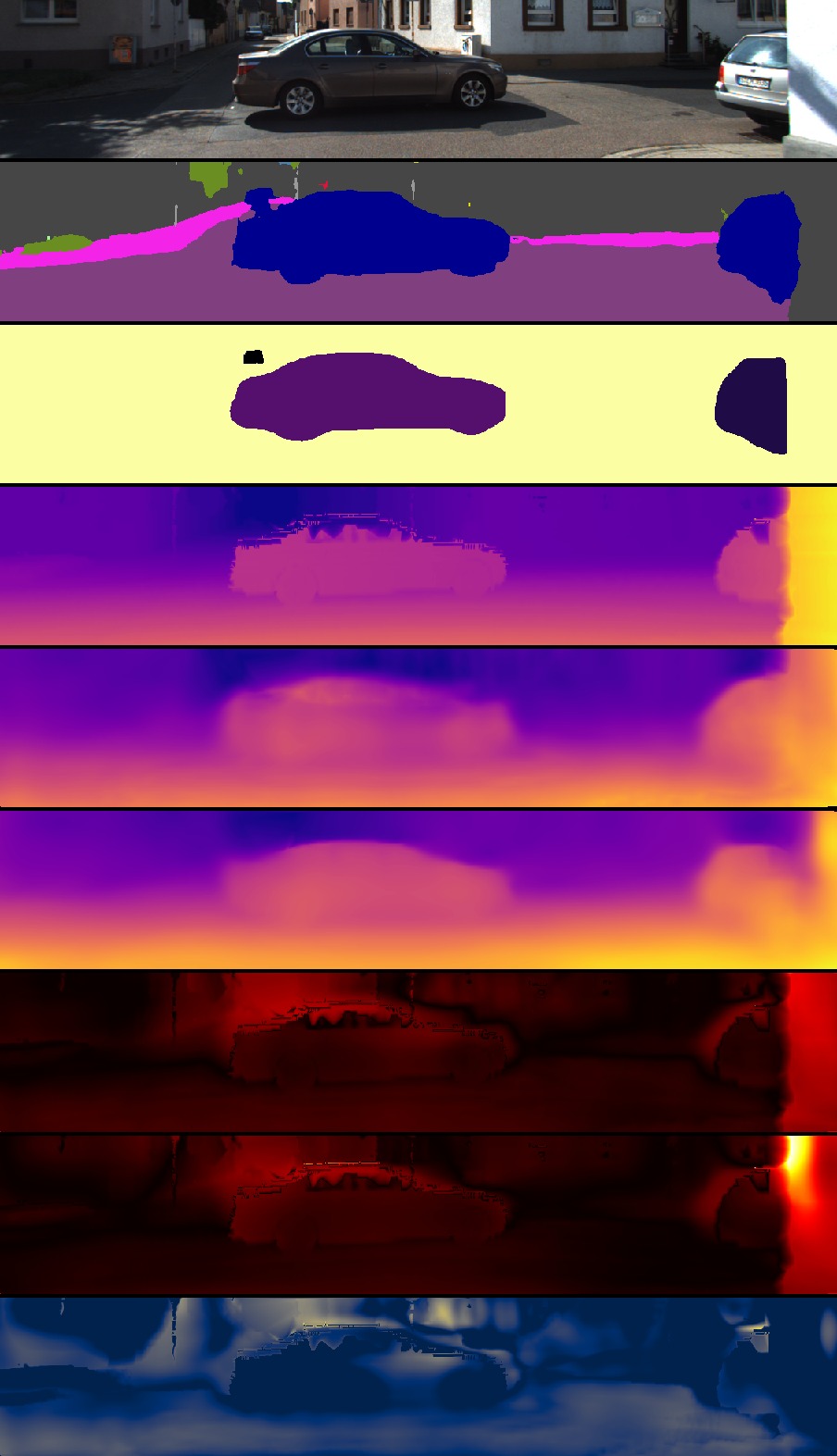}
     \end{subfigure}
    \caption{\CAPOUR}
\end{figure}

\begin{figure}[!hbtp]
    \centering
         \begin{subfigure}[b]{0.48\linewidth}
      \includegraphics[width=\linewidth]{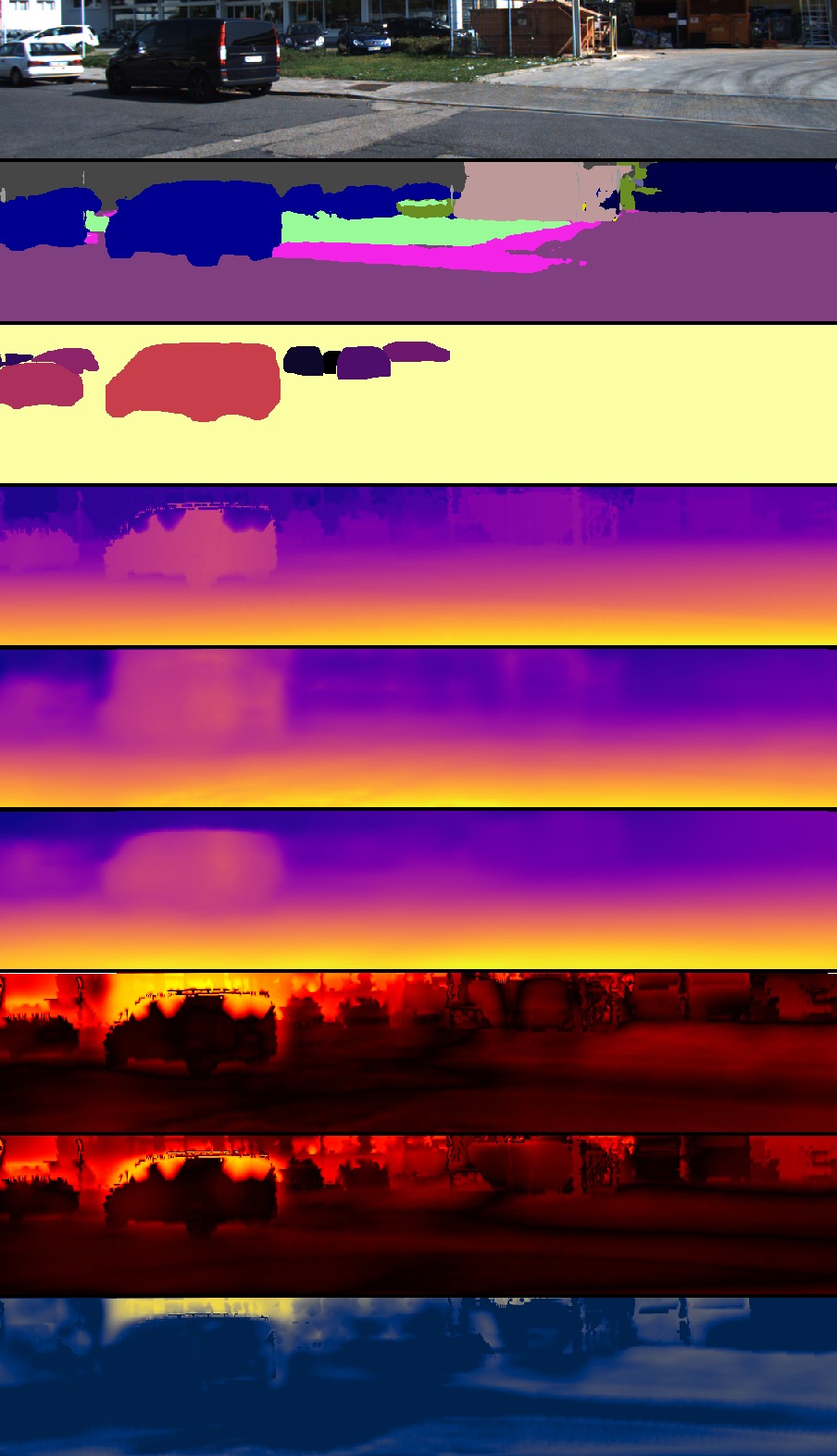}
     \end{subfigure}
     \GAP
     \begin{subfigure}[b]{0.48\linewidth}
      \includegraphics[width=\linewidth]{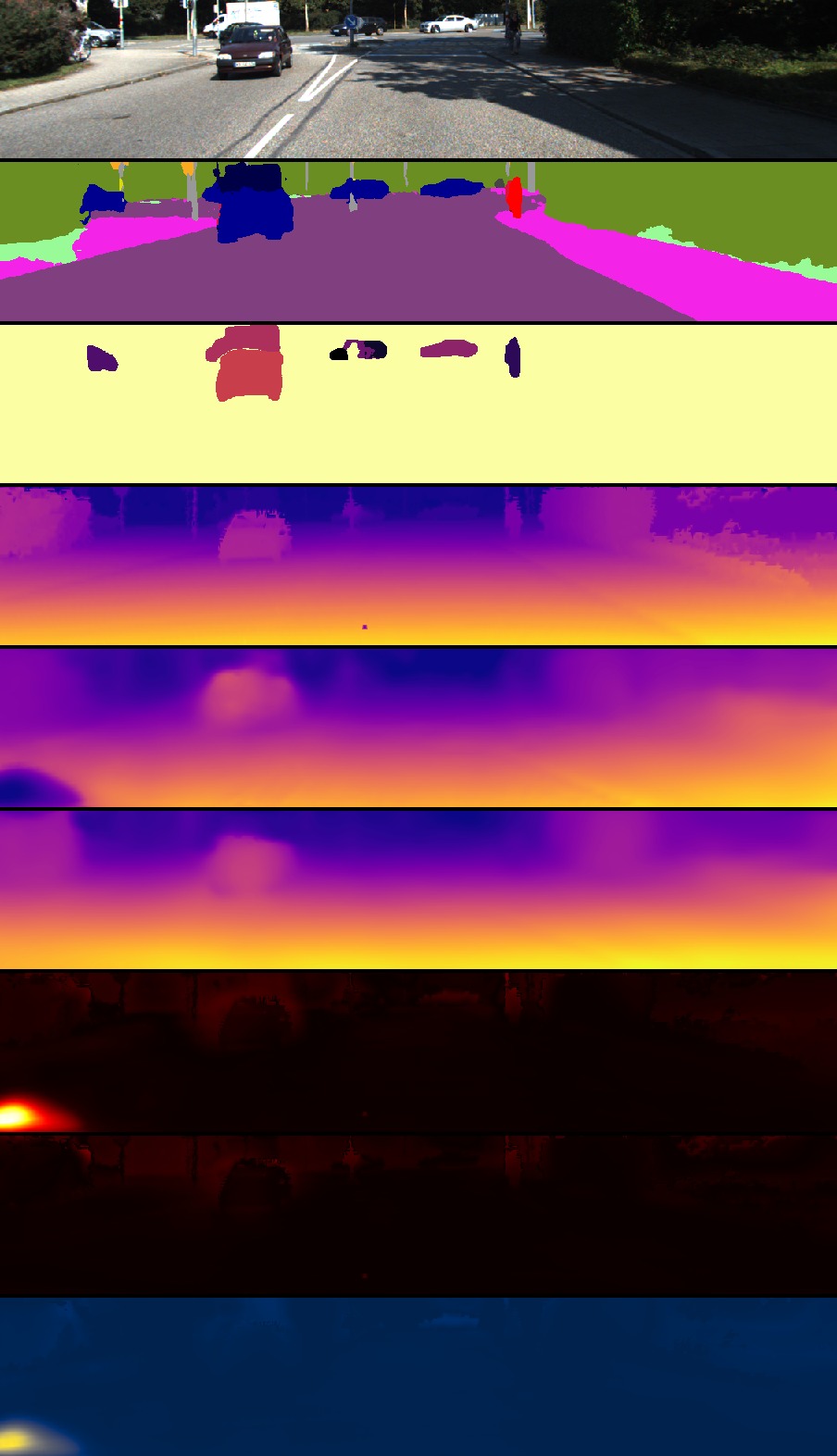}
     \end{subfigure}
    \caption{\CAPOUR}
\end{figure}

\begin{figure}[!hbtp]
    \centering
    \begin{subfigure}[b]{0.48\linewidth}
      \includegraphics[width=\linewidth]{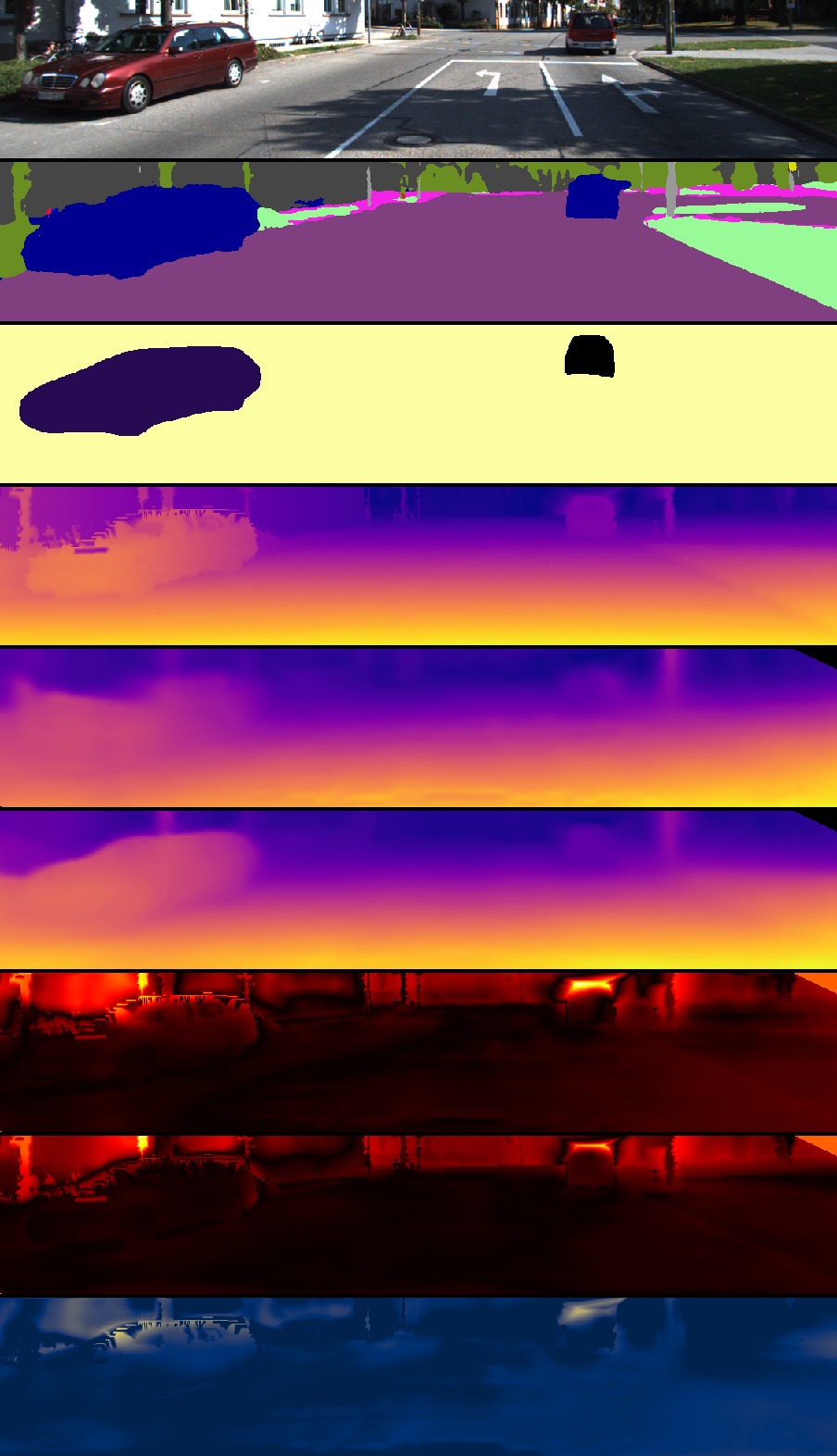}
     \end{subfigure}
     \GAP
         \begin{subfigure}[b]{0.48\linewidth}
      \includegraphics[width=\linewidth]{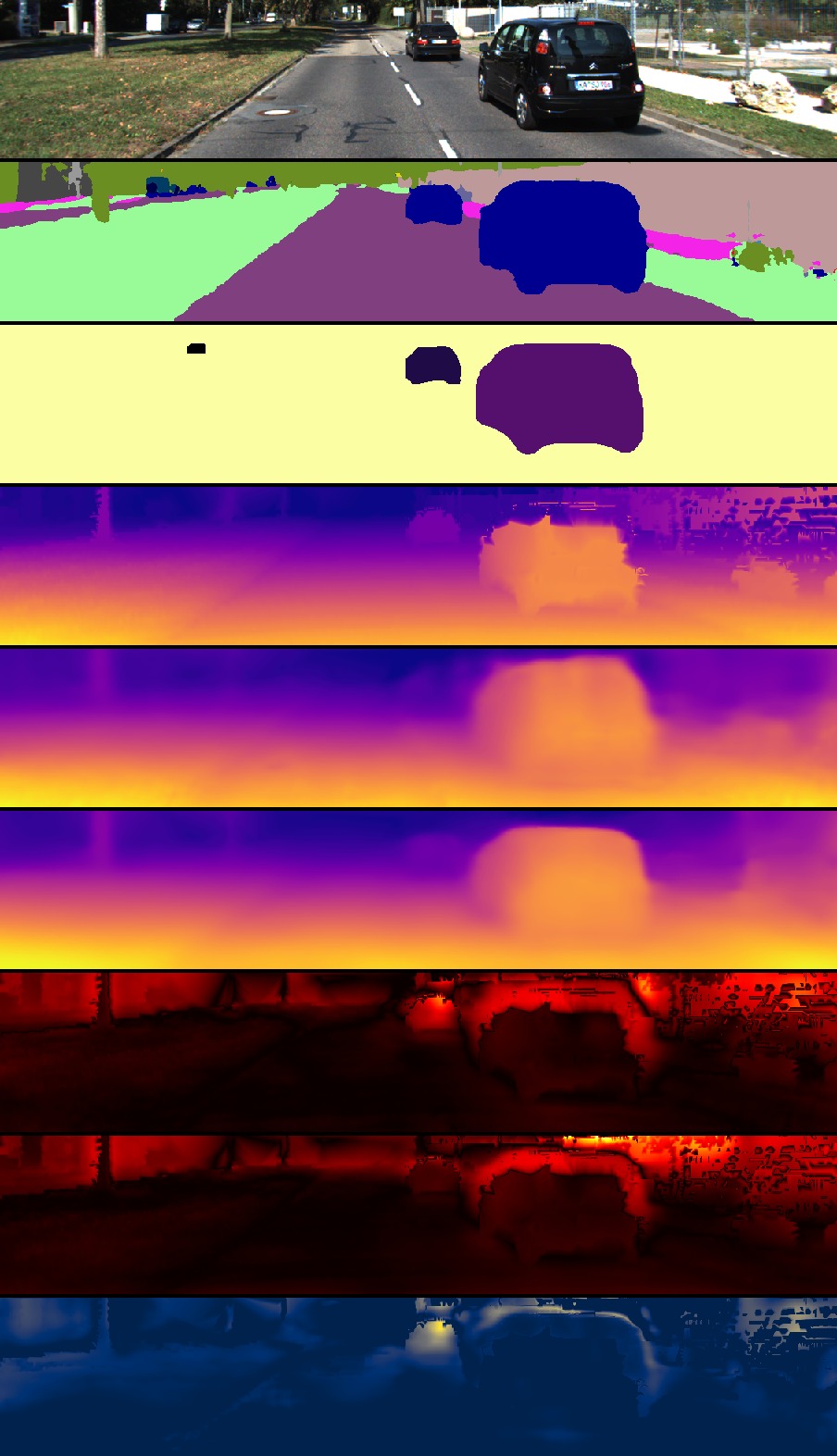}
     \end{subfigure}
    \caption{\CAPOUR}
\end{figure}

\end{document}